\documentclass{article}

\PassOptionsToPackage{numbers}{natbib}
\usepackage[main, final]{neurips_2026}
\usepackage[utf8]{inputenc} % allow utf-8 input
\usepackage[T1]{fontenc}    % use 8-bit T1 fonts
\usepackage{hyperref}       % hyperlinks
\usepackage{url}            % simple URL typesetting
\usepackage{booktabs}       % professional-quality tables
\usepackage{amsfonts}       % blackboard math symbols
\usepackage{nicefrac}       % compact symbols for 1/2, etc.
\usepackage{microtype}      % microtypography
\usepackage{xcolor}         % colors
\usepackage{graphicx}
\usepackage{float}
\usepackage{algorithm}
\usepackage[noend]{algorithmic}
\usepackage{graphicx}%
\usepackage{multirow}%
\usepackage{amsmath}
\usepackage{gensymb}
\usepackage{booktabs}
\usepackage{xcolor}
\usepackage{colortbl}
\usepackage{tabularx}
\definecolor{bestcolor}{RGB}{204, 229, 204}

\title{Structure-Semantic Co-optimized Latent Diffusion Model for Fast Visual Anagram Synthesis}

\author{%
 Xiang Gao, \ \ Yunpeng Jia\thanks{Corresponding author.}\\
 School of Digital Media and Design Arts\\
 Beijing University of Posts and Telecommunications\\
 \texttt{\{gaoxiang1102, xibei156\}@bupt.edu.cn} \\
}
\begin{document}

\maketitle

\begin{abstract}
  Visual anagram is an intriguing form of art creation wherein a single image presents different conceptual interpretations under transformations such as flipping or rotation. Recent work has achieved visual anagram synthesis by leveraging pretrained text-to-image (T2I) diffusion models, yet still suffers from several key limitations including computational inefficiency, suboptimal aesthetic quality, and weak semantic fidelity and expressiveness. This work focuses on generating visual anagrams with substantially improved visual quality at minimal computational cost, thereby advancing intelligent creation of illusionary digital art. To increase image resolution while reducing time overhead, we adapt the cutting-edge parallel denoising algorithm from pixel-based T2I model to the adversarially distilled latent-based one, and accordingly propose a structure-semantic co-optimization (S2CO) framework to counteract the consequent visual degradation. As the core of our approach, S2CO framework comprises three key innovations: (\romannumeral1) null-text structure alignment optimization; (\romannumeral2) semantic enhancement optimization; (\romannumeral3) attention-guided noise fusion. Building upon these components, our method dubbed \textbf{S2CO-Anagram} is able to generate higher-resolution anagram images with noticeably superior visual harmony, aesthetic quality, and semantic faithfulness to the text prompt than related SOTA approaches, all while achieving substantially faster inference speed. 
\end{abstract}

\section{Introduction}
Visual illusions occur when human perception deviates from physical reality. For decades, this intriguing phenomenon has captivated researchers across psychology, neuroscience, and computer vision. Recently, computationally synthesizing visual illusions has blossomed into a compelling area, encompassing algorithms that span classical optimization to modern deep generative models.

Early parametric approaches formulate visual camouflage \cite{chu2010camouflage,owens2014camouflaging,zhang2020deep}, hybrid perception \cite{oliva2006hybrid}, geometric illusion \cite{ehm2011variational,chi2014optical}, motion illusion \cite{freeman1991motion}, and color illusion \cite{ulucan2024computational} as tractable optimization problems. These approaches typically require per-instance optimization and are limited in generalization ability.

With the proliferation of deep learning, GANs \cite{goodfellow2020generative} achieve remarkable success in synthesizing illusions \cite{gomez2022synthesis} and camouflages \cite{guo2023ganmouflage}, marking a paradigm shift from hand-crafted parametric models to learnable generative models. More recently, techniques have leveraged generative prior of pretrained T2I diffusion models to synthesize optical illusion hidden images via score distillation \cite{burgert2024diffusion}, noise decomposition \cite{geng2024factorized}, and phase manipulation \cite{gao2025ptdiffusion}. Practical applications fuse functional data with artistic expression, yielding scannable QR codes that double as aesthetic illusions \cite{liao2025diffqrcoder,liao2024diffusion}. Collectively, these developments mark a new era in computational visual illusion synthesis. 

Among the various forms of optical illusions, visual anagrams, images that transform into distinct percepts upon specific transformations, has garnered increasing attention recently. Diffusion Illusions \cite{burgert2024diffusion} optimizes an image based on the generative prior of T2I diffusion model, ensuring that its transformed views semantically align with different text prompts. However, this method yields limited resolution and quality, and the reliance on score distillation \cite{poole2022dreamfusion} optimization leads to prohibitive time consumption. SyncTweedies \cite{kim2024synctweedies} builds a multi-view synchronous denoising architecture, achieving zero-shot anagram generation by fusing the predicted clean images across views at each DDIM \cite{song2020denoising} denoising step. Visual Anagrams \cite{geng2024visual}, in contrast, achieves the same goal by averaging the multi-view noise predictions during the denoising process. Anagram-MTL \cite{xu2025diffusion} improves upon Visual Anagrams by introducing attention-based anti-segregation optimization and CLIP-based noise vector balancing, mitigating issues of ``concept segregation'' and ``concept domination'' respectively. However, these three methods rely on pixel-based T2I models, resulting in high computational overhead and limited image resolution. LookingGlass \cite{chang2025lookingglass} realizes visual anagram generation based on Laplacian pyramid warping. This method achieves high image resolution by building upon latent-based diffusion model, though the repeated image warping and aggregation operations still incur prolonged inference time.

\begin{figure}[t]
	\centering
	\includegraphics[width=0.99\textwidth]{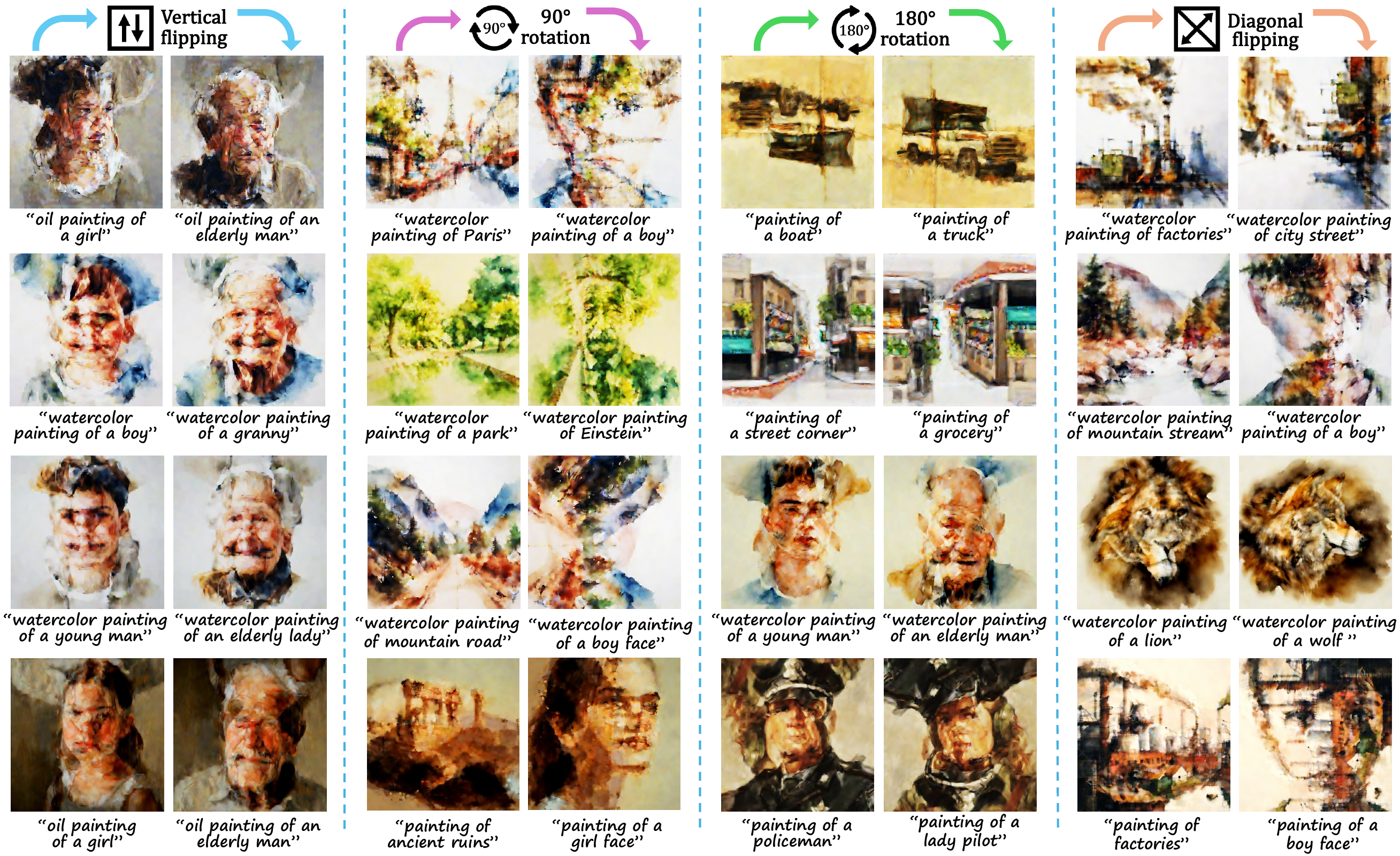}
	\caption{Our method enables fast synthesis of high-quality visual anagram images that conceptually conform to different text prompts after undergoing a transformation such as flipping and rotation, offering artists and designers an efficient tool to create optical illusion visual arts.}
	\label{fig:teaser}
\end{figure}

In this paper, we propose S2CO-Anagram, a lightweight latent diffusion model (LDM) for fast and high-quality visual anagram synthesis with example results displayed in Fig. \ref{fig:teaser}. To achieve high resolution and inference speed, S2CO-Anagram inherits the multi-view parallel denoising paradigm \cite{kim2024synctweedies,geng2024visual,xu2025diffusion} and builds upon SDXL-Turbo \cite{citekey}—an adversarially distilled \cite{sauer2024adversarial} LDM enabling high-resolution and few-step T2I synthesis. However, directly applying multi-view synchronous denoising to LDM fails to generate visually harmonious and semantically faithful results, a challenge we attribute to the following two problems. \textbf{(1) Structural Divergence}: the denoising of multi-view latents structurally diverges under different semantic guidance. For instance, "horse" and "penguin" differ inherently in shape and pose, driving their prompt-guided denoising processes toward different structures. However, averaging their structural signals via multi-view noise fusion forces the latent into a compromised structure that cannot faithfully capture either concept. Moreover, cross-view structural conflicts can lead to rigid, unnatural content fusion, producing less harmonious visual anagrams. These issues are more pronounced in latent-based T2I models especially those requiring few denoising steps. \textbf{(2) Semantic Dilution}: the multi-view latents are denoised toward different semantic directions. Averaging the predicted noises inevitably dilutes semantic information, undermining conceptual faithfulness of the generated result to both text prompts. Consequently, we propose a structure-semantic co-optimization (S2CO) framework to tackle the aforementioned two problems.

To tackle the degradation in semantic fidelity and visual harmony caused by structural divergence, we propose \textbf{null-text structure alignment optimization}, which penalizes structural deviation at each denoising step by optimizing the null-text embedding in classifier-free guidance (CFG) with our proposed \textbf{View-Cycle Low-Frequency Consistency (VLC) Loss}. The VLC loss aligns spatial structure of the predicted clean latents across views in the DCT (Discrete Cosine Transform) domain, suppressing structural divergence along the synchronous denoising trajectory and thereby remarkably mitigating the corresponding issues of semantic vagueness and structural artifacts.

To mitigate semantic dilution problem caused by noise averaging, we propose semantic enhancement optimization and attention-guided noise fusion. The former approach explicitly enhances semantic distinctiveness by optimizing the input latent at each denoising step to maximally activate vision-text cross-attention maps associated to key concepts. The latter one leverages attention maps to guide spatially adaptive noise fusion, which preserves semantically salient features of both views more faithfully than simple noise averaging, and thereby alleviating semantic blurring.

To conclude, the key contributions of this work are summarized as follows:
\begin{itemize}
	\item We propose S2CO-Anagram, a lightweight few-step LDM that efficiently generates high-quality visual anagrams via a novel structure-semantic co-optimization algorithm.
	\item A novel null-text structure alignment optimization is proposed to suppress structural divergence along the multi-view synchronous denoising process, promoting semantic faithfulness and visual harmony of the generated anagram images from a frequency-domain perspective.
	\item A semantic enhancement optimization which enhances conceptual fidelity and discernibility of the generated results by elevating cross-attention activations associated with key concepts.
	\item An attention-guided noise fusion scheme is proposed to promote semantic saliency and distinctiveness by leveraging cross-attention maps to guide spatially adaptive noise fusion.
	\item Our S2CO-Anagram outperforms related approaches both qualitatively and quantitatively in visual anagram generation, all while achieving substantially faster inference speed.
\end{itemize}

\section{Related Work}
\subsection{Diffusion models and applications}. 
Since the advent of DDPM \cite{ho2020denoising}, diffusion models have soon emerged as the mainstream in generative modeling \cite{dhariwal2021diffusion}, and have achieved rapid advancements in inference speed \cite{songdenoising,lu2022dpm}, conditional generation \cite{saharia2022palette,batzolis2021conditional,yu2023freedom}, structural control \cite{zhang2023adding,mou2024t2i,cho2025att}, and concept customization \cite{lin2024non,zhang2024attention,jiang2025infiniteyou}. With LDM \cite{rombach2022high} dramatically lowing computational overhead by transferring diffusion model from high-dimentional pixel space to low-dimensional feature space, a plethora of visual applications built upon LDM have sprung up, such as text-guided image editing \cite{mokady2023null,nguyen2025swiftedit,cao2023masactrl,zhang2025enabling}, text-driven image-to-image (I2I) translation \cite{tumanyan2023plug,gao2024frequency,gao2024fbsdiff,gao2026fbsdiff++}, image colorization \cite{zabari2023diffusing,liang2025control}, image inpainting \cite{corneanu2024latentpaint,liu2024structure}, and image restoration \cite{xia2023diffir,fei2023generative,zhu2023denoising}. This work focus on leveraging off-the-shelf T2I diffusion model for high-quality and efficient visual anagram synthesis.
\subsection{AI-driven visual art creation}. 
The intersection of AI and artistic creation has witnessed significant advancements in recent years. The seminal work of Neural Style Transfer \cite{gatys2016image} achieved stunning success in separating and recombining image content and style using convolutional neural network, reproducing artistic styles of renowned paintings on natural images. The algorithm was later substantially advanced in real-time efficiency \cite{johnson2016perceptual,huang2017arbitrary,li2017universal}, arbitrary-style generalization \cite{park2019arbitrary,liu2021adaattn,zhang2022domain}, and improved visual quality \cite{lin2021drafting,chen2021artistic,kolkin2019style}, spawning a wide range of artistic creation applications such as fashion design \cite{jiang2021deep}, text stylization \cite{fu2018style}, and decorated logo generation \cite{atarsaikhan2018contained}. With the growing prevalence of GANs, a broader spectrum of visual art creation has emerged within the paradigm of cross-domain I2I translation \cite{isola2017image,zhu2017unpaired}, spanning applications such as artistic painting stylization \cite{yi2019apdrawinggan,gao2020rpd,gao2025sragan}, image cartoonization \cite{gao2022learning,jiang2023scenimefy}, and artistic font generation \cite{xie2021dg,yang2019controllable}. Under the more recent diffusion-based paradigm, researchers have explored adapting large-scale T2I diffusion models for artistic style transfer \cite{zhang2023inversion,chung2024style,fahim2025stam} or style-specific content creation \cite{hertz2024style,zhou2025attention,aravanis2025only}, elevating the quality and controllability of artistic creation to an unprecedented level. Beyond image-based artistic creation, recent advances have extended artistic stylization into the 3D domain \cite{fang20253,zhang2025stylizedgs,mei2024regs}, achieving reference-based stylization of arbitrary 3DGS \cite{kerbl20233d} scenes and thereby enabling artistic expression on immersive, freely navigable scenes. Further examples of diffusion-based visual art generation include aesthetic QR code creation \cite{liao2025diffqrcoder}, optical illusion hidden image generation \cite{gao2025ptdiffusion}, design image generation \cite{wang2025designdiffusion}, artistic typography generation \cite{liu2025arttypo}, and visual anagram synthesis \cite{geng2024visual}. This work focuses on generating visual anagrams with substantially improved visual quality at minimal inference time, thereby pushing the frontiers in the creation of illusionary digital art.

\section{Proposed Method}
This section first clarifies the overall model architecture, then elaborates on the key components of our method, and finally concludes with the full algorithm and implementation details. The preliminary backgrounds of diffusion model and classifier-free guidance (CFG) are described in the \textbf{Appendix}.
\subsection{Overall Model Architecture}
\begin{figure*}[t]
	\centering
	\includegraphics[width=0.99\linewidth]{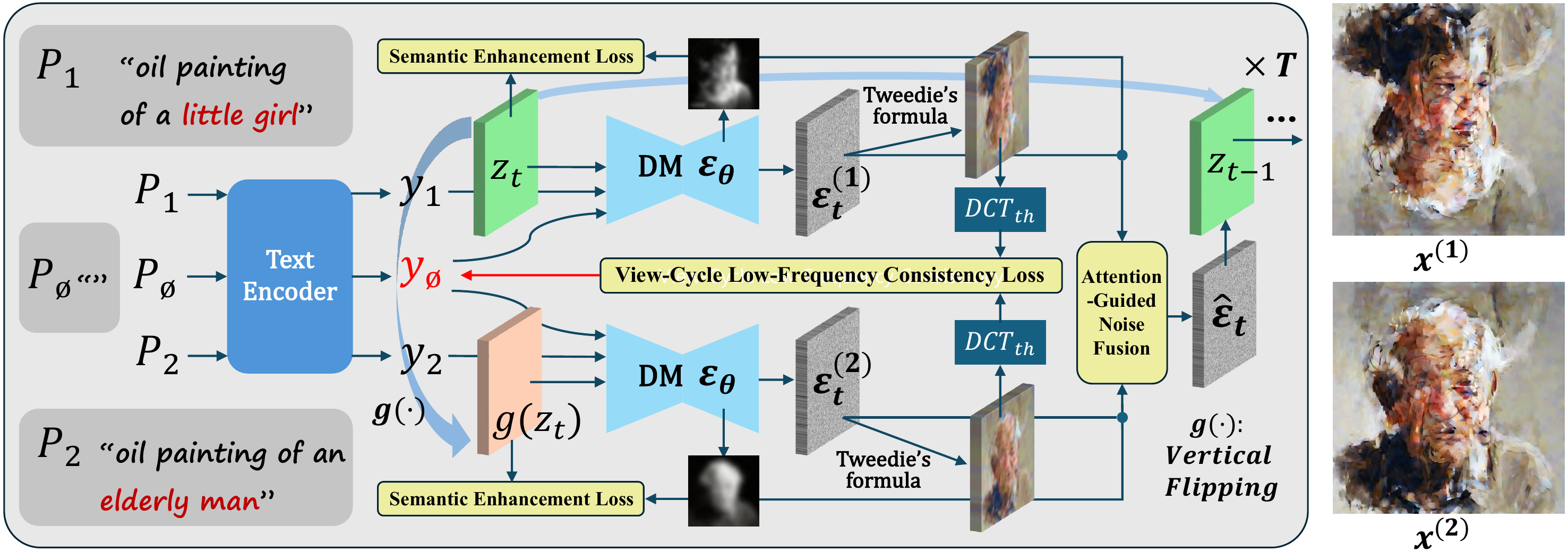}
	\caption{\textbf{Method overview}. Our method synchronously denoises two latent views linked by a specified transformation followed by fusing their noise estimations at each denoising step. Grounding on this process, a novel optimization framework comprising null-text structure alignment, semantic enhancement, and attention-guided noise fusion is proposed to enhance the generated results.}
	\label{fig:method}
\end{figure*}

Given a pair of text prompts ($P_{1}, P_{2}$) and a geometric transformation $g(\cdot)$, our method synthesizes a pair of anagram images ($x^{(1)},x^{(2)}$), where $x^{(1)}$ and $x^{(2)}$ are semantically aligned with $P_{1}$ and $P_{2}$, respectively, and satisfy $x^{(2)}=g(x^{(1)})$, $x^{(1)}=g^{-1}(x^{(2)})$.

The overall architecture of S2CO-Anagram is illustrated in Fig. \ref{fig:method}. Our model is built upon pretrained LDM. The denoising network $\epsilon_{\theta}$ is conditioned on the text embeddings ($y_{1}, y_{2}$) which are encoded from ($P_{1}, P_{2}$), the null-text embedding $y_{\emptyset}$ encoded from $P_{\emptyset}$=``'' is employed for CFG. At each denoising step $t$, the current denoising latent $z_{t}$ and its transformed view $g(z_{t})$ are input to $\epsilon_{\theta}$ along with their paired text embeddings, yielding the predicted noises $\epsilon_{t}^{(1)}$ and $\epsilon_{t}^{(2)}$, respectively:
\begin{equation}
	\epsilon_{t}^{(1)}=\omega\epsilon_{\theta}(z_{t}, t, y_{1})+(1-\omega)\epsilon_{\theta}(z_{t}, t, y_{\emptyset}), \ \
	\epsilon_{t}^{(2)}=\omega\epsilon_{\theta}(g(z_{t}), t, y_{2})+(1-\omega)\epsilon_{\theta}(g(z_{t}), t, y_{\emptyset}).
	\label{eq:noise_pred}
\end{equation}
Where $\omega$ is the guidance strength in CFG \cite{ho2021classifier}. The predicted noises $\epsilon_{t}^{(1)}$ and $\epsilon_{t}^{(2)}$ are integrated via the attention-guided noise fusion module, yielding the fused noise $\hat{\epsilon}_{t}$ that incorporates semantic concepts of both $P_{1}$ and $P_{2}$. Then, the next-step denoising latent $z_{t-1}$ is computed and iterated as:
\begin{equation}
	z_{t-1}=\frac{\sqrt{\bar{\alpha}_{t-1}}(z_{t}-\sqrt{1-\bar{\alpha}_{t}}\hat{\epsilon}_{t})}{\sqrt{\bar{\alpha}_{t}}} + \sqrt{1-\bar{\alpha}_{t-1}}\hat{\epsilon}_{t},
	\label{eq:next_latent}
\end{equation}
where $\{\bar{\alpha}_{t}\}$ are DDPM \cite{ho2020denoising} schedule parameters. The multi-view parallel denoising is repeated for $T$ steps to yield the noise-free clean latent $z_{0}$. Let $D$ denote the LDM VAE decoder, the final pair of generated visual anagram images are obtained as:
\begin{equation}
	x^{(1)}= \frac{D(z_{0})+g^{-1}(D(g(z_{0})))}{2}, \ \
	x^{(2)}= \frac{D(g(z_{0}))+g(D(z_{0}))}{2}.
	\label{eq:final_anagram}
\end{equation}
The VAE decoder $D$ is not transformation-invariant, i.e., $g(D(z_{0})) \neq D(g(z_{0}))$, the averaging-based formulation of Eq. \ref{eq:final_anagram} is designed to maximize the transformation consistency between $x^{(1)}$ and $x^{(2)}$. 

\subsection{Null-Text Structure Alignment}
Structural divergence occurs when the multi-view latents are denoised under different prompt guidance, that is, $g^{-1}(\epsilon_{t}^{(2)})$ tends to structurally deviate from $\epsilon_{t}^{(1)}$ to cater to different semantic concepts. Such structural misalignment between multi-view noise estimations yields a compromised structure in the fused noise $\hat{\epsilon}_{t}$, which degrades semantic faithfulness of the resulting $z_{t-1}$ to both $y_{1}$ and $y_{2}$, and can lead to unnatural structural artifacts in the final results. Hence, we propose an optimization strategy to tackle such structural divergence issue. Since structural information is imperceptible in the noise space, we apply Tweedie's formula so as to align structure based on the predicted clean latents:
\begin{equation}
	z_{t\rightarrow0}^{(1)}=\frac{z_{t}-\sqrt{1-\bar{\alpha}_{t}}\epsilon_{t}^{(1)}}{\sqrt{\bar{\alpha}_{t}}}, \ \  z_{t\rightarrow0}^{(2)}=\frac{g(z_{t})-\sqrt{1-\bar{\alpha}_{t}}\epsilon_{t}^{(2)}}{\sqrt{\bar{\alpha}_{t}}}.
	\label{eq:tweedie}
\end{equation}
To enforce structural consistency between $z_{t\rightarrow0}^{(1)}$ and $g^{-1}(z_{t\rightarrow0}^{(2)})$, we apply latent-space DCT transformation to constrain their low-frequency consistency, considering that DCT low-frequency spectral components correspond to global structure and layout. Let $dct_{th}^{(h)}(\cdot)$ denote the function of applying 1D DCT transformation over the $h$ dimension and extracting the frequency components lower than the percentile threshold $th$,  $dct_{th}^{(w)}(\cdot)$ be the correspondingly similar function applied over the $w$ dimension (with details of these two functions illustrated and clarified in the \textbf{Appendix}). We design View-Cycle Low-Frequency Consistency (VLC) loss which enforces low-frequency consistency between the predicted clean latents $z_{t\rightarrow0}^{(1)}$ and $g^{-1}(z_{t\rightarrow0}^{(2)})$ for structural alignment:
\begin{equation}
		L_{VLC}=\left|\left|dct_{th}^{(h)}(z_{t\rightarrow0}^{(1)})-dct_{th}^{(h)}(g^{-1}(z_{t\rightarrow0}^{(2)}))\right|\right|_{2}^{2}+ 
		\left|\left|dct_{th}^{(w)}(z_{t\rightarrow0}^{(1)})-dct_{th}^{(w)}(g^{-1}(z_{t\rightarrow0}^{(2)}))\right|\right|_{2}^{2}.
		\label{eq:VLC}
\end{equation}
To maximally disentangle structural alignment with semantic guidance, we minimize $L_{VLC}$ w.r.t. the null-text embedding $y_{\emptyset}$. That is, $y_{1}$ and $y_{2}$ define the multi-view semantic concepts of the generated result while $y_{\emptyset}$ is employed for per-step structural alignment optimization, which is formulated as:
\begin{equation}
	y_{\emptyset}:=y_{\emptyset}-\lambda_{1}\nabla_{y_{\emptyset}}L_{VLC},
	\label{eq:vlc_optimize}
\end{equation}
where $\lambda_{1}$ is the initial learning rate. By enforcing the two views to share a common low-frequency structure, The distinguishing features between the two concepts (e.g., dog vs. cat) are pushed into high-frequency details. After DCT alignment optimization, the resulting two noise predictions exhibit less structural conflict, leading to less semantic compromise and more harmonious content fusion.

\subsection{Cross-Modal Semantic Enhancement}
Fusing the multi-view noise predictions $\epsilon_{t}^{(1)}$ and $\epsilon_{t}^{(2)}$ into $\hat{\epsilon}_{t}$ integrates semantic concepts of each view, while also diluting them. To compensate for such semantic dilution issue, we propose cross-modal semantic enhancement optimization which tunes the input latent of $\epsilon_{\theta}$ at each denoising step to maximize cross-attention activations associated to key concepts of $P_{1}$ and $P_{2}$. Cross-attention maps capture correlation intensity between textual semantics and spatial locations, which are computed as:
\begin{equation}
	M_{l}^{(1)}=Softmax\left(\frac{Q_{l}^{(1)} (K_{l}^{(1)})^{T}}{ \sqrt{d}}\right), \ \ \
	M_{l}^{(2)}=Softmax\left(\frac{Q_{l}^{(2)} (K_{l}^{(2)})^{T}}{ \sqrt{d}}\right),
\end{equation}
where $l$ is layer index, $M_{l}^{(1)}$ and $M_{l}^{(2)}$ denote attention maps at the $l$-th cross-attention layer inside $\epsilon_{\theta}$ given $z_{t}$ and $g(z_{t})$ as input, respectively, $Q_{l}^{(1)}$ and $Q_{l}^{(2)}$ are queries of the $l$-th cross-attention layer derived from $z_{t}$ and $g(z_{t})$, respectively, $K_{l}^{(1)}$ and $K_{l}^{(2)}$ are keys of the $l$-th cross-attention layer originated from $y_{1}$ and $y_{2}$, respectively. Based on all $\{M_{l}^{(1)}\}$ and $\{M_{l}^{(2)}\}$, attention maps corresponding to conceptually most crucial words in $P_{1}$ and $P_{2}$ (marked in red in Fig. \ref{fig:method}) are extracted, normalized, and summed to yield the final $\tilde{M}^{(1)}$ and $\tilde{M}^{(2)}$ (with details described in the \textbf{Appendix}) which capture correlation intensity of each spatial location to the key concept of $P_{1}$ and $P_{2}$, respectively. To compensate for semantic dilution issue, we enhance semantic expressiveness by maximizing pixel activation of $\tilde{M}^{(1)}$ and $\tilde{M}^{(2)}$ w.r.t. the input latent $z_{t}$, which is formulated as:

\begin{equation}
	z_{t} := z_{t} + \lambda_{2} \nabla_{z_{t}} \left( \sum\nolimits_{(i,j) \in \Omega^{(1)}} \tilde{M}^{(1)}_{i,j} + \sum\nolimits_{(i,j) \in \Omega^{(2)}} \tilde{M}^{(2)}_{i,j} \right),
	\Omega^{(k)}=\left\{(i,j)|\tilde{M}^{(k)}_{i,j}\geq \tau_{p}^{(k)}\right\},
	\label{eq:se_optimize}
\end{equation}
where $i,j$ denote spatial indices, $\lambda_{2}$ is the initial learning rate, $\tau_{p}^{(k)}$ refers to the $p$-th percentile of the values in $\tilde{M}^{(k)}$. To avoid indiscriminate semantic enhancement that leads to semantic bleeding to all spatial locations, we only back-propagate through locations above the $p$-th percentile of each attention map, thereby reinforcing semantic expressiveness only in the most concept-relevant regions.
\subsection{Attention-Guided Noise Fusion}
Observing that direct noise fusion via averaging tends to produce blurring and semantically ambiguous results, which we attribute to the lack of spatial awareness of salient semantic concept, as average noise fusion applies equal-strength weighting to all locations. Therefore, we propose to leverage the concept saliency spatial indication provided by the extracted attention maps $\tilde{M}^{(1)}$ and $\tilde{M}^{(2)}$ for spatially adaptive noise fusion, which is formulated as:
\begin{equation}
	\hat{\epsilon}_{t} = \epsilon_{t}^{(1)} \cdot \frac{\tilde{M}^{(1)}}{\tilde{M}^{(1)} + g^{-1}(\tilde{M}^{(2)})}+g^{-1}(\epsilon_{t}^{(2)}) \cdot \frac{g^{-1}(\tilde{M}^{(2)})}{\tilde{M}^{(1)}+g^{-1}(\tilde{M}^{(2)})}.
	\label{eq:adaptive_noise_fusion}
\end{equation}

\begin{algorithm}[t]
	\caption{Complete algorithm of S2CO-Anagram}
	\label{algorithm}
	\begin{algorithmic}[1]
		\renewcommand{\algorithmicrequire}{\textbf{Input:}}
		\renewcommand{\algorithmicensure}{\textbf{Output:}}
		\REQUIRE{A transformation $g(\cdot)$, source view text prompt $P_{1}$, transformed view text prompt $P_{2}$.}
		\ENSURE{Generated anagram image pair ($x^{(1)}, x^{(2)}$).}
		\STATE Randomly initialize a noise latent $z_{T} \sim \mathcal{N}(0, I)$.
		\FOR{$t=T$ to $1$}
		\STATE Initialize a new null-text embedding $y_\emptyset{}$ encoded from the null-text prompt $P_{\emptyset}$=``''.
		\FOR{$k=1$ to $K$}
		\STATE Input $z_{t}$ and $g(z_{t})$ to $\epsilon_{\theta}$, obtaining ($\epsilon_{t}^{(1)}$, $\epsilon_{t}^{(2)}$) and computing ($z_{t\rightarrow0}^{(1)}$, $z_{t\rightarrow0}^{(2)}$), ($\tilde{M}^{(1)}$, $\tilde{M}^{(2)}$).
		\STATE Update $y_{\emptyset}$ with a step of null-text structure alignment optimization using Eq. \ref{eq:vlc_optimize}.
		\STATE Update $z_{t}$ with a step of cross-modal semantic enhancement optimization using Eq. \ref{eq:se_optimize}.
		\ENDFOR
		\STATE Recompute noise predictions ($\epsilon_{t}^{(1)}$, $\epsilon_{t}^{(2)}$) with the optimized $y_{\emptyset}$ and $z_{t}$ using Eq. \ref{eq:noise_pred}.
		\STATE Obtain the fused noise $\hat{\epsilon}_{t}$ via Eq. \ref{eq:adaptive_noise_fusion}, based on which compute $z_{t-1}$ via Eq. \ref{eq:next_latent}.
		\ENDFOR 
		\STATE Obtain the clean latent $z_{0}$ and generate the final anagram image pair ($x^{(1)}$, $x^{(2)}$) using Eq. \ref{eq:final_anagram}.
	\end{algorithmic}
\end{algorithm}

\subsection{Structure-Semantic Co-optimization}
Integrating the above three key components, we propose a structure-semantic co-optimized LDM for high-quality and fast visual anagram synthesis. At each denoising step, null-text structural alignment and semantic enhancement are jointly carried out prior to the attention-guided noise fusion. The complete algorithm of S2CO-Anagram is shown in Alg. \ref{algorithm}. During per-step optimization, we set $\lambda_{1}$=0.1 in Eq. \ref{eq:vlc_optimize}, $\lambda_{2}$=0.5 in Eq. \ref{eq:se_optimize}, and set the DCT low-pass percentile threshold $th$=5 in Eq. \ref{eq:VLC}. We adopt the few-step LDM SDXL-Turbo as the T2I backbone to maximize inference speed, with the total number of denoising steps set to $T$=4 and the CFG guidance strength to $\omega$=1.5. As shown in Algorithm \ref{algorithm}, a $K$-step inner loop is applied at each denoising step for structure-semantic co-optimization, with $K$=5 in our implementation. The percentile parameter in Eq. \ref{eq:se_optimize} is set $p$=60. Though the per-step optimization introduces additional computational overhead, the few-step denoising of SDXL-Turbo still keeps the overall time cost remarkably low, achieving considerably faster inference speed than related approaches. Our method demonstrates that high-quality visual anagram synthesis is also attainable using latent-based T2I diffusion models, even the distilled few-step ones.

\begin{figure*}[t]
	\centering
	\includegraphics[width=0.99\linewidth]{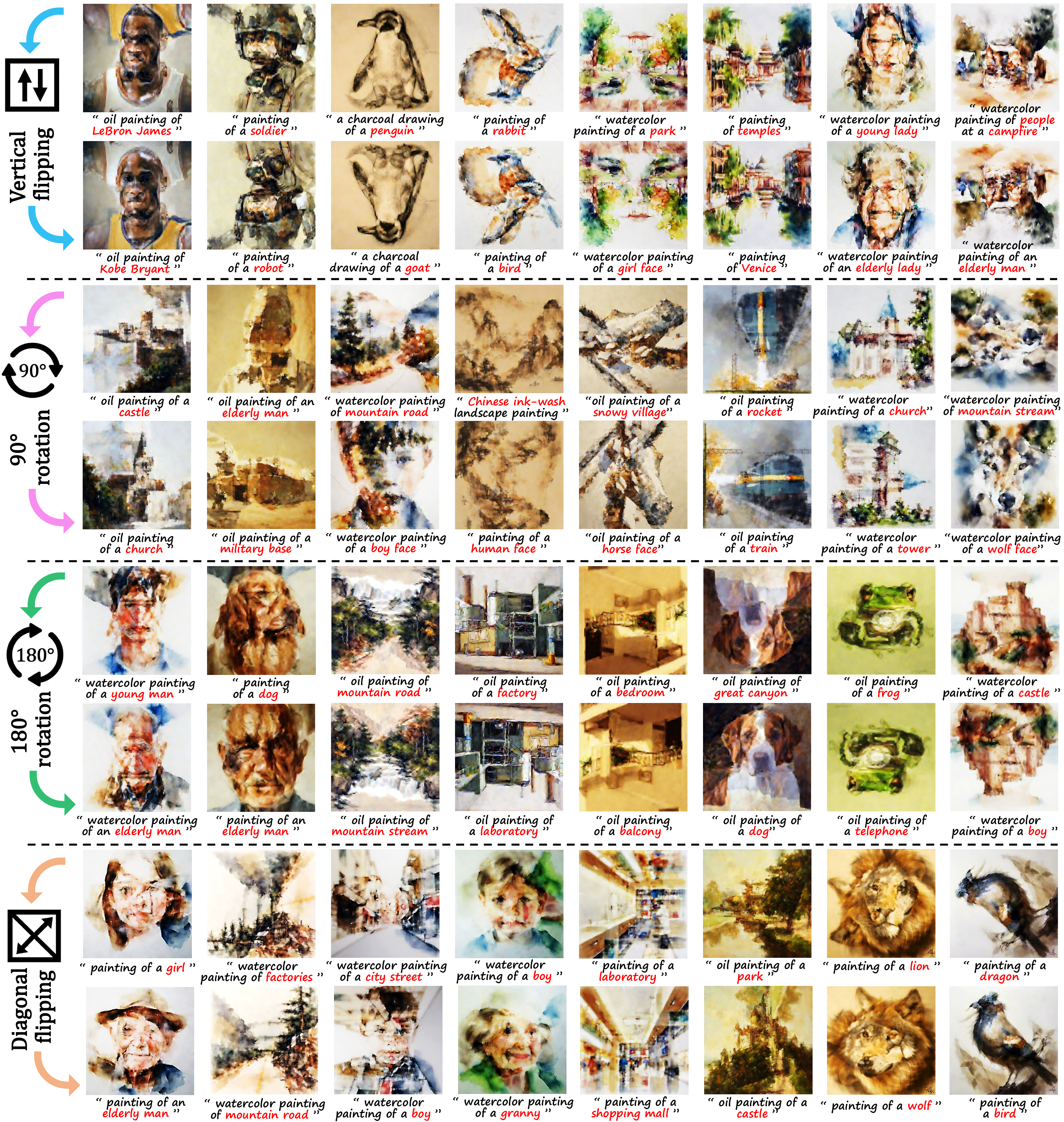}
	\caption{Visual anagram synthesis results of our S2CO-Anagram using different transformation functions $g(\cdot)$ with keywords in each text prompt highlighted in \textcolor{red}{red}. \textbf{Better viewed with zoom in}.}
	\label{fig:qualitative_res}
\end{figure*}

\begin{figure*}[t]
	\centering
	\includegraphics[width=0.99\linewidth]{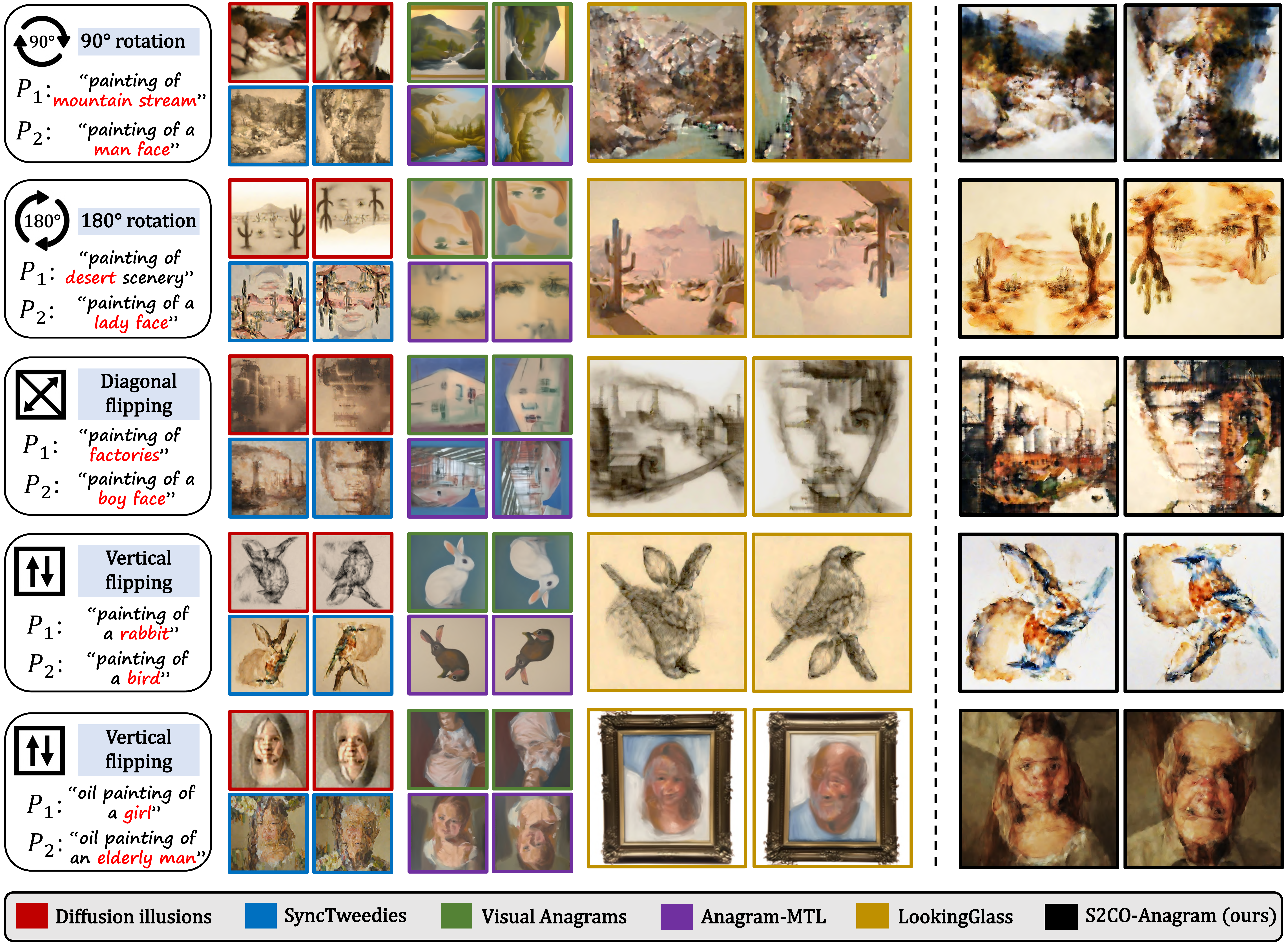}
	\caption{Qualitative method comparison. Results are presented at their actual resolution: thumbnails correspond to 256×256 outputs, and larger panels correspond to 512×512 results.}
	\label{fig:method_compare}
\end{figure*}

\section{Experimental Results}
\subsection{Qualitative Evaluations}
\noindent \textbf{Qualitative results and comparisons}. As showcased in Fig. \ref{fig:qualitative_res},  our method produces high-quality anagram images under different transformations such as vertical flipping, diagonal flipping, 90\degree \ rotation, 180\degree \ rotation. The generated results are semantically faithful to both $P_{1}$ and $P_{2}$ in their corresponding views. The visual concepts of the two views are fused in a harmonious and tactful manner, allowing each view to retain its structural integrity and semantic clarity without sacrificing those of the other. In Fig. \ref{fig:method_compare}, we qualitatively compare our method with related SOTA approaches in synthesizing visual anagrams. Among the compared methods, Diffusion Illusions \cite{burgert2024diffusion} exhibits relatively weak semantic fidelity, where key concepts from text prompts are not sufficiently and clearly expressed, which we attribute to the instability, high-variance gradient estimates, and mode collapse inherent in the score distillation optimization.  SyncTweedies \cite{kim2024synctweedies} generates results that are plagued by non-trivial noise and artifacts. Its fusion of concepts appears overly rigid, compromising visual harmony across both views. This suggests that multi-view fusion in feature space yields less smooth results and is more prone to structural artifacts compared to noise-space fusion. Visual Anagrams \cite{geng2024visual} and Anagram-MTL \cite{xu2025diffusion} produce results with limited semantic clarity, where key concepts from text prompts are not sufficiently discernible. Unlike the aforementioned methods based on image-space T2I model that are constrained to low-resolution outputs (256$\times$256), LookingGlass \cite{chang2025lookingglass} synthesizes higher-resolution anagram images by leveraging latent-based model. Nevertheless, its dependence on image-space warping operations introduces a lack of smoothness and naturalness into the synthesized results. In contrast, our S2CO-Anagram qualitatively outperforms all the above approaches in image resolution, conceptual fidelity, semantic clarity, and visual harmony. 
\begin{figure*}[t]
	\centering
	\includegraphics[width=0.99\linewidth]{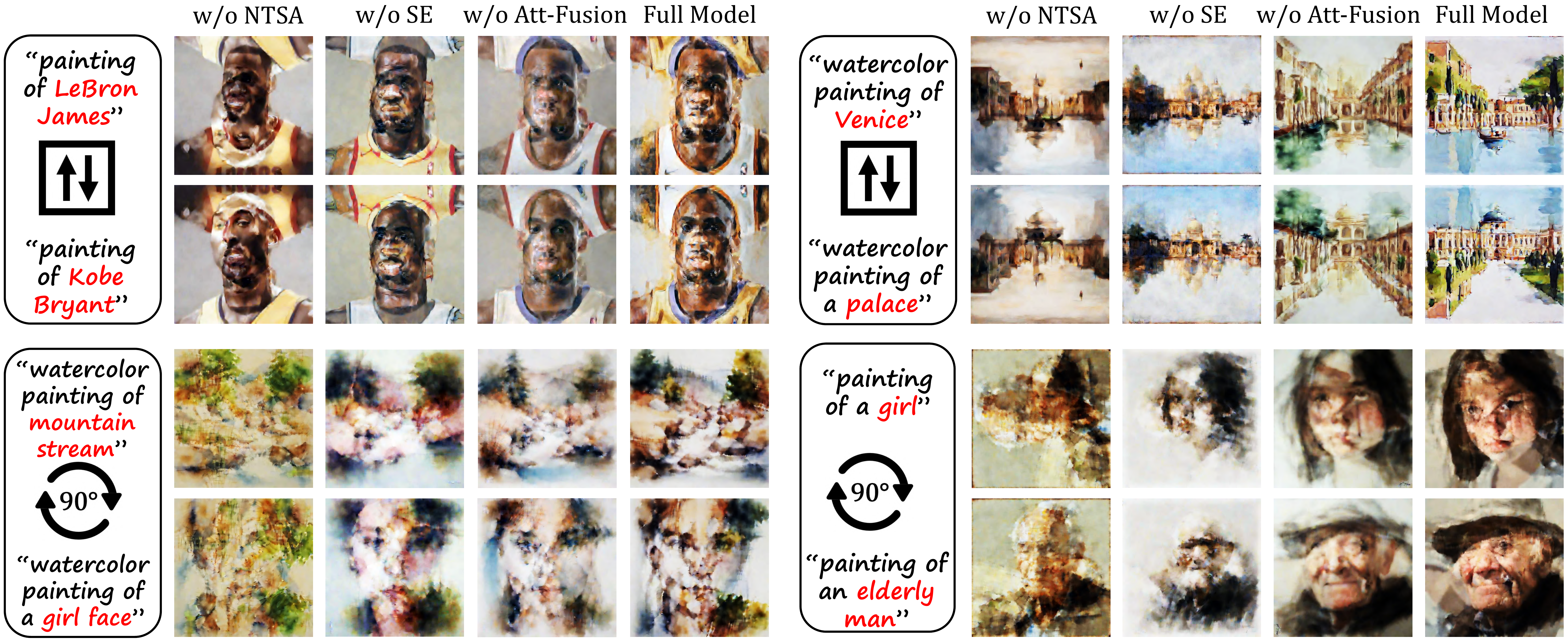}
	\caption{Qualitative ablation studies on key components of our method including null-text structure alignment (NTSA), semantic enhancement (SE), and attention-guided noise fusion (Att-Fusion).}
	\label{fig:ablation_study}
\end{figure*}

\begin{figure*}[t]
	\centering
	\includegraphics[width=0.99\linewidth]{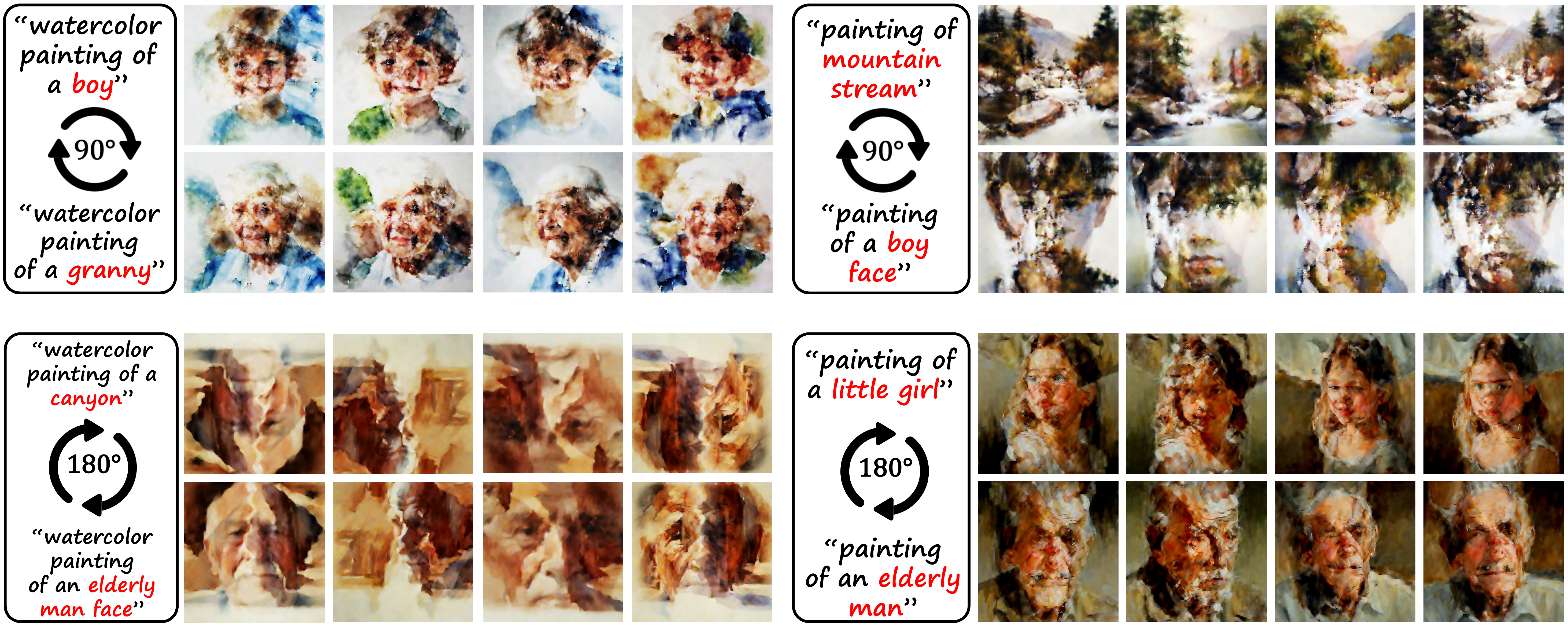}
	\caption{Diverse results with fixed text prompts ($P_{1},P_{2}$) is enabled by varying the noise latent $z_{T}$.}
	\label{fig:diversity}
\end{figure*}

\begin{figure*}[!bt]
	\centering
	\includegraphics[width=0.99\linewidth]{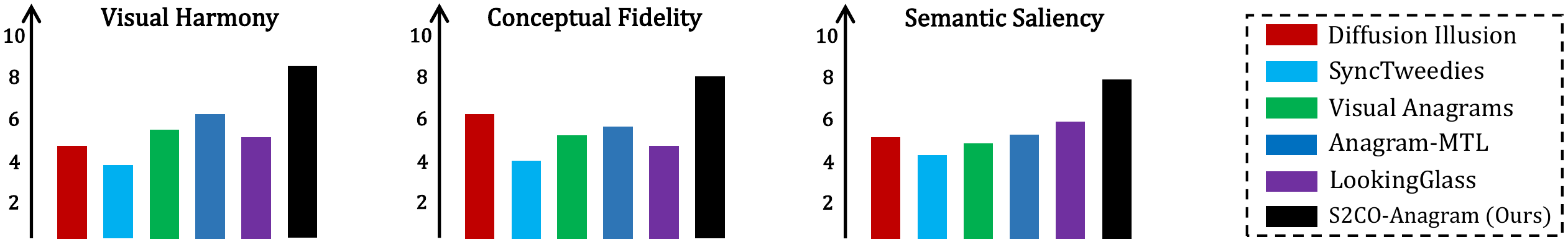}
	\caption{A user study comparing different methods for generating anagram images.}
	\label{fig:user_score}
\end{figure*}

\textbf{Qualitative ablation study}. To assess contribution of each core component in our method, a qualitative ablation study is performed and displayed in Fig. \ref{fig:ablation_study}. Results w/o null-text structure alignment (NTSA) exhibit severe structural artifacts, where the two views fail to fuse harmoniously. This is because the structural divergence along the multi-view denoising process pushes the fused latent toward a compromised structure which undermines semantic fidelity to both $P_{1}$ and $P_{2}$. By narrowing structural deviation via NTSA, the synchronous denoising converges toward a less conflicting direction, producing results that are visually more harmonious and conceptually more faithful. Compared with results w/o semantic enhancement (SE) optimization, the full model produces outputs that more clearly express the target semantics, demonstrating effectiveness of SE in promoting semantic fidelity. Results w/o attention-guided noise fusion (Att-Fusion) are relatively blurred and less detailed than those of the full model, demonstrating advantage of Att-Fusion over average-based noise fusion in preserving salient and fine-grained semantic features. 

\textbf{Generation diversity}. As demonstrated in Fig. \ref{fig:diversity}, our method enables random synthesis of diverse anagram images under fixed text prompts ($P_{1}, P_{2}$) by varying the initial Gaussian noise latent $z_{t}$. When $z_{t}$ remains fixed, the deterministic nature of our sampling process yields a unique result.

\begin{table}[t]
	\centering
	\small
	\setlength{\extrarowheight}{0pt}
	\renewcommand{\arraystretch}{1}
	\setlength{\abovetopsep}{0pt}
	\setlength{\belowbottomsep}{0pt}
	\caption{Quantitative evaluations across different methods in visual anagram synthesis.}
	\label{tab:quantitative_evaluations}
	\resizebox{\textwidth}{!}{%
	\begin{tabular}{lcccccc}
		\toprule
		\textbf{Method} & $\mathcal{A}_{min}\uparrow$ & $\mathcal{C}\uparrow$ & $\mathcal{A}_{avg}\uparrow$ & $\mathcal{AS}\uparrow$ & Resolution $\uparrow$ & Inference Time $\downarrow$ \\
		\midrule
		Diffusion Illusions & 0.2547 & 0.6867 & 0.2676 & 4.46 & 256$\times$256 & 1027.3s \\
		SyncTweedies       & 0.2568 & 0.6896 & 0.2692 & 4.28 & 256$\times$256 & 5.9s   \\
		Visual Anagrams    & 0.2583 & 0.6744 & 0.2717 & 5.37 & 256$\times$256 & 4.0s   \\
		Anagram-MTL        & 0.2711 & 0.6913 & 0.2816 & 5.65 & 256$\times$256 & 16.2s   \\
		LookingGlass       & 0.2574 & 0.6848 & 0.2703 & 4.96 & 512$\times$512 & 74.8s   \\
		\textbf{S2CO-Anagram} & \cellcolor{bestcolor}{\textbf{0.2726}} & \cellcolor{bestcolor}{\textbf{0.6918}} & \cellcolor{bestcolor}{\textbf{0.2824}} & \cellcolor{bestcolor}{\textbf{6.14}} & \cellcolor{bestcolor}{\textbf{512$\times$512}} & \cellcolor{bestcolor}{\textbf{2.6s}} \\
		\bottomrule
	\end{tabular}
	}
\end{table}

\begin{table}[t]
	\centering
	\begin{minipage}{0.58\textwidth}
		\centering
		\small
		\caption{Quantitative ablation study on the contributions of the key model components.}
		\label{tab:ablation_study}
		\renewcommand{\arraystretch}{1}
		\setlength{\tabcolsep}{2pt}
		\resizebox{\textwidth}{!}{%
		\begin{tabular}{lcccccc}
			\toprule
			\textbf{NTSA} & \textbf{SE} & \textbf{Att-Fusion} & $\mathcal{A}_{min}\uparrow$ & $\mathcal{C}\uparrow$ & $\mathcal{A}_{avg}\uparrow$ & $\mathcal{AS}\uparrow$\\
			\midrule
			$\times$ & $\times$ & $\times$ & 0.2463 & 0.6672 & 0.2658 & 4.64\\
			$\checkmark$ & $\times$ & $\times$ & 0.2703 & 0.6894 & 0.2805 & 5.58\\
			$\times$ & $\checkmark$ & $\times$ & 0.2577 & 0.6745 & 0.2738 & 5.12\\
			$\times$ & $\times$ & $\checkmark$ & 0.2482 & 0.6690 & 0.2677 & 4.83\\
			$\times$ & $\checkmark$ & $\checkmark$ & 0.2603 & 0.6786 & 0.2764 & 5.29\\
			$\checkmark$ & $\times$ & $\checkmark$ & 0.2710 & 0.6903 & 0.2813 & 5.84\\
			$\checkmark$ & $\checkmark$ & $\times$ & 0.2719 & 0.6914 & 0.2819 & 6.06\\
			$\checkmark$ & $\checkmark$ & $\checkmark$ & \cellcolor{bestcolor}{\textbf{0.2726}} & \cellcolor{bestcolor}{\textbf{0.6918}} & \cellcolor{bestcolor}{\textbf{0.2824}} & \cellcolor{bestcolor}{\textbf{6.14}}\\
			\bottomrule
		\end{tabular}}
	\end{minipage}
	\hfill
	\begin{minipage}{0.4\textwidth}
		\centering
		\small
		\renewcommand{\arraystretch}{1}
		\setlength{\tabcolsep}{4pt}
		\caption{Ablation study on the DCT low-pass filtering percentile threshold $th$.}
		\label{tab:ablation_th}
		\resizebox{\textwidth}{!}{%
		\begin{tabular}{lccc}
			\toprule
			$DCT_{th}$ & $\mathcal{A}_{min}\uparrow$ & $\mathcal{C}\uparrow$ & $\mathcal{A}_{avg}\uparrow$ \\
			\midrule
			$DCT_{0}$  & 0.2603 & 0.6786 & 0.2764 \\
			$DCT_{1}$  & 0.2655 & 0.6839 & 0.2785 \\
			$DCT_{2}$  & 0.2684 & 0.6887 & 0.2797 \\
			$DCT_{3}$  & 0.2710 & 0.6906 & 0.2811 \\
			$DCT_{4}$  & 0.2722 & 0.6916 & 0.2821 \\
			$DCT_{5}$  & \cellcolor{bestcolor}{\textbf{0.2726}} & \cellcolor{bestcolor}{\textbf{0.6918}} & \cellcolor{bestcolor}{\textbf{0.2824}} \\
			$DCT_{10}$ & 0.2699 & 0.6895 & 0.2803 \\
			$DCT_{20}$ & 0.2616 & 0.6793 & 0.2772 \\
			\bottomrule
		\end{tabular}}
	\end{minipage}
\end{table}

\subsection{Quantitative Analyses}
\textbf{Quantitative evaluations}. We quantitatively evaluate methods on CIFAR10 2-views \cite{xu2025diffusion}, a dataset with text prompt pairs using 10 classes of objects from CIFAR-10 \cite{krizhevsky2009learning}. For each pair of text prompts, we generate 10 images with different seeds and report the average results. We use CLIP \cite{radford2021learning} to measure visual-text semantic consistency (text fidelity) in the following three dimensions: (1) \textbf{Worst Alignment Score ($\mathcal{A}_{min}$)}: the worst image-text CLIP score across two views; (2) \textbf{Concealment Score ($\mathcal{C}$)}: measuring how well the generated anagram image from one view conceals textual concepts from the other view; (3) \textbf{Average Alignment Score ($\mathcal{A}_{avg}$)}: the average image-text CLIP score across two views. The mathematical definitions of these metrics are defined and clarified in the \textbf{Appendix}. Besides, we measure visual quality of the generated results by evaluating the average \textbf{Aesthetic Score} ($\mathcal{AS}$) via the pretrained LAION Aesthetics Predictor V2 \cite{schuhmann_improved_aesthetic}. Finally, we evaluate model inference time which are averaged across 5 samples on a single RTX PRO 6000 GPU. Among the compared methods, our S2CO-Anagram delivers superior image resolution with significantly lower inference time, and consistently achieves the best performance in text fidelity metrics ($\mathcal{A}_{min}$, $C$, $\mathcal{A}_{avg}$) and visual aesthetic metric $\mathcal{AS}$, indicating that our method generates the most visually appealing anagram images with the clearest semantic discrimination and the highest computational efficiency.

\textbf{Quantitative ablation study}. Quantitative ablation study on core components of our method is reported in Tab. \ref{tab:ablation_study}. Results show that all key contributions of our method synergetically improve the performance, which is basically in line with the qualitative results shown in Fig. \ref{fig:ablation_study}. Among the proposed contributions, NTSA plays the most crucial role in promoting semantic fidelity of the generated anagram images. Ablation study on the DCT low-pass filtering percentile threshold $th$ in NTSA is reported in Tab. \ref{tab:ablation_th}. The absence of NTSA ($th$=0) yields the worst metrics, increasing $th$ leads to strengthened structure alignment, thereby improving performance by alleviating structural artifacts. However, increasing $th$ beyond an optimal point imposes excessive constraint on cross-view structure consistency, which is detrimental to the precise semantic expressiveness of the target concepts, resulting in degraded performance. We set $th$=5 based on empirical parameter tuning.

\textbf{User study}. We conduct a user study for subjective evaluation of the compared methods. Over 40 participants were recruited to rate the generated anagram images of our model and related methods on a 1–10 scale, with evaluation details provided in the Appendix. Fig. \ref{fig:user_score} presents the average user ratings for all compared methods. Our S2CO-Anagram surpasses other approaches by a large margin across all the evaluated dimensions, demonstrating significant advantages of our method for synthesizing visually harmonious, aesthetic appealing, and semantically faithful visual anagrams.

\section{Conclusion}
This paper presents S2CO-Anagram, a latent diffusion model (LDM) for fast and high-quality visual anagram synthesis. Tailored for the structural divergence and semantic dilution issues of the multi-view parallel denoising, we propose a structure-semantic co-optimization (S2CO) framework comprising null-text structure alignment, semantic enhancement, and attention-guided noise fusion. By integrating S2CO with pretrained few-step LDM, our method is able to generate higher-resolution anagram images with noticeably superior visual harmony, conceptual fidelity, semantic clarity, and aesthetic quality than related approaches, all while achieving significantly faster inference speed. %However, our method still suffers from two limitations: it struggles to generate realistic-style anagram images despite producing visually appealing artistic visual anagrams, and it fails to yield satisfying results for more complex transformation functions such as jigsaw rearrangement. We will explore solutions to these two limitations in future work.

\section{Acknowledgements}
This work was supported by the National Natural Science Foundation of China [Grant No. 6260076518], and was also supported by the Major Project of the National Social Science Fund of China (Art Studies) [Grant No. 26ZD12].

\newpage
\appendix

\section{Preliminary Backgrounds}
\subsection{Diffusion Model Background}
Since the advent of the Denoising Diffusion Probabilistic Model (DDPM) \cite{ho2020denoising}, diffusion models have soon dominated the field of generative AI, owing to their advantages in training stability and sampling diversity compared to GANs. Grounded in the theory of stochastic differential equations, a diffusion model learns to iteratively denoise a noise-corrupted input signal (e.g., an image or video clip), ultimately generating clean data that follow the target distribution. Conceptually, the model comprises a forward diffusion process and a reverse denoising process. The forward process gradually adds noise to the data over a series of steps, transforming the data into a standard Gaussian distribution. The reverse process learns to invert the forward process by iteratively removing noise, starting from pure noise and progressively reconstructing the original data. The model is trained to predict the noise added at each step of the forward process. Through this denoising objective, the model can generate new data samples by initializing from random noise and applying the reverse process.

Given the original data distribution $q(x_{0})$, the forward diffusion process applies a $T$-step Markov chain to gradually add noise to the original data $x_{0}$ according to the conditional distribution $q(x_{t}|x_{t-1})$, which is defined as:
\begin{equation}
	q(x_{t}|x_{t-1})=\mathcal{N}(x_{t}; \sqrt{\alpha_{t}}x_{t-1}, (1-\alpha_{t})\mathcal{I}),
\end{equation}
where $\alpha_{t}$ follows a predefined schedule, $\alpha_{t}\in(0, 1)$, $\alpha_{t} > \alpha_{t+1}$. Using the notation $\bar{\alpha}_{t}=\prod_{i=1}^{t}\alpha_{i}$, we can derive the marginal distribution $q(x_{t}|x_{0})$ that can be used to directly obtain $x_{t}$ from $x_{0}$ in a single step for arbitrary time step $t$:
\begin{equation}
	q(x_{t}|x_{0})=\mathcal{N}(x_{t}; \sqrt{\bar{\alpha}_{t}}x_{0}, (1-\bar{\alpha}_{t})\mathcal{I}),
	\label{xt|x0}
\end{equation}
where $\sqrt{\bar{\alpha}_{T}} \approx 0$. With the forward diffusion process, the source data $x_{0}$ is transformed into $x_{T}$ that follows an isotropic Gaussian distribution.

The reverse denoising process learns to conversely convert a Gaussian noise $x_{T}$ to the manifold of the original data distribution $q(x_{0})$ by gradually estimating and sampling from the posterior distribution $p(x_{t-1}|x_{t})$. Since the posterior distribution $p(x_{t-1}|x_{t})$ is mathematically intractable, we can derive the conditional posterior distribution $p(x_{t-1}|x_{t}, x_{0})$ with the Bayes formula and some algebraic manipulation:
\begin{equation}
	p(x_{t-1}|x_{t},x_{0})=\mathcal{N}(x_{t-1}; \tilde{\mu}_{t}(x_{t}, x_{0}), \tilde{\beta}_{t}\mathcal{I}),
	\label{conditional_posterior}
\end{equation}
\begin{equation}
	\tilde{\mu}_{t}(x_{t}, x_{0})=\frac{\sqrt{\bar{\alpha}_{t-1}}\beta_{t}}{1-\bar{\alpha}_{t}}x_{0}+\frac{\sqrt{\alpha}_{t}(1-\bar{\alpha}_{t-1})}{1-\bar{\alpha}_{t}}x_{t},
\end{equation}
\begin{equation}
	\tilde{\beta}_{t}=\frac{1-\bar{\alpha}_{t-1}}{1-\bar{\alpha}_{t}}\beta_{t},
\end{equation}
where $\beta_{t}=1-\alpha_{t}$. 
However, the conditional posterior distribution $p(x_{t-1}|x_{t}, x_{0})$ cannot be directly used for sampling since $x_{0}$ is unavailable at inference time ($x_{0}$ is the target of the sampling process). Therefore, DDPM tries to estimate the unknown $x_{0}$ given the $x_{t}$ at each time step. Considering the reparameterization form of Eq. \ref{xt|x0}:
\begin{equation}
	x_{t}=\sqrt{\bar{\alpha}_{t}}x_{0}+\sqrt{1-\bar{\alpha}_{t}}\epsilon_{t},
	\label{reparam}
\end{equation}
where $\epsilon_{t}$ denotes the randomly sampled Gaussian noise that maps $x_{0}$ to $x_{t}$ in a single step according to Eq. \ref{xt|x0}. Given Eq. \ref{reparam}, we can represent $x_{0}$ using $x_{t}$ and $\epsilon_{t}$:
\begin{equation}
	x_{0}=\frac{1}{\sqrt{\bar{\alpha}_{t}}}(x_{t}-\sqrt{1-\bar{\alpha}_{t}}\epsilon_{t}).
	\label{repre_x0_with_xt}
\end{equation}

However, the Gaussian noise $\epsilon_{t}$ sampled in the forward diffusion process is also unknown for the reverse denoising process where only $x_{t}$ is available. Consequently, DDPM builds a noise estimation network $\epsilon_{\theta}$ that predicts the sampled Gaussian noise $\epsilon_{t}$ in Eq. \ref{repre_x0_with_xt} with $x_{t}$ and time step $t$ as input, which is realized by training $\epsilon_{\theta}$ with the following noise regression loss:
\begin{equation}
	L=\|\epsilon_{t}-\epsilon_{\theta}(x_{t}, t)\|_{2},
	\label{DDPM_loss}
\end{equation}
where $t\sim$ Uniform($\{1,...,T\}$), $\epsilon_{t} \sim \mathcal{N}(0, \mathcal{I})$, $x_{t}$ is computed via Eq. \ref{reparam}. After model training, $y_{\theta}(x_{t})$, the estimation of $x_{0}$ given $x_{t}$, can be obtained simply by replacing $\epsilon_{t}$ in Eq. \ref{repre_x0_with_xt} with the predicted noise $\epsilon_{\theta}(x_{t}, t)$:
\begin{equation}
	y_{\theta}(x_{t})=\frac{1}{\sqrt{\bar{\alpha}_{t}}}(x_{t}-\sqrt{1-\bar{\alpha}_{t}}\epsilon_{\theta}(x_{t}, t)).
	\label{x_0_pred}
\end{equation}

Replacing the unknown $x_{0}$ in Eq. \ref{conditional_posterior} with its predicted estimation $y_{\theta}(x_{t})$ given by Eq. \ref{x_0_pred}, we can sample $x_{t-1}$ based on $x_{t}$ from the approximate posterior distribution $\mathcal{N}(x_{t-1};\tilde{\mu}_{t}(x_{t},y_{\theta}(x_{t})),\tilde{\beta}_{t}\mathcal{I})$, and thus sample the ultimate $x_{0}$ step by step from the initial Gaussian noise $x_{T}$. 

\subsection{Conditional Diffusion Model}
Taking the image generation task as an example, conditional diffusion model tackles conditional image synthesis by introducing additional condition $c$ to the model to guide image generation (denoising) process. In this paradigm, the condition signal $c$ together with $x_{t}$ and time step $t$ are taken as input to the noise estimation network $\epsilon_{\theta}$, such that $\epsilon_{\theta}$ is trained to conditionally predict the added Gaussian noise in the forward diffusion process, as supervised by the randomly sampled $\epsilon_{t}$ in Eq. \ref{reparam}. The training loss given by Eq. \ref{DDPM_loss} is correspondingly updated as:
\begin{equation}
	L=\|\epsilon_{t}-\epsilon_{\theta}(x_{t}, t, c)\|_{2},
	\label{conditional_DDPM_loss}
\end{equation}
where $t\sim$ Uniform($\{1,...,T\}$), $\epsilon_{t} \sim \mathcal{N}(0, \mathcal{I})$, $x_{t}$ is computed via Eq. \ref{reparam}. After model training, the reverse sampling process is applied to generate new images from random Gaussian noise $x_{T}$, based on the step-by-step denoising according to the conditional posterior distribution given by Eq. \ref{conditional_posterior}, in which the unknown $x_{0}$ is approximated by the linear combination of $x_{t}$ and the conditional noise estimation, \textit{i.e.}, the $y_{\theta}(x_{t})$ (approximate $x_{0}$ estimated by $x_{t}$) in Eq. \ref{x_0_pred} is updated as:
\begin{equation}
	y_{\theta}(x_{t}, c)=\frac{1}{\sqrt{\bar{\alpha}_{t}}}(x_{t}-\sqrt{1-\bar{\alpha}_{t}}\epsilon_{\theta}(x_{t}, t, c)).
\end{equation}
For text-to-image (T2I) diffusion model, the condition signal $c$ is the textual embedding encoded from the input text prompt.

\subsection{Latent Diffusion Model}
Latent diffusion model (LDM) \cite{rombach2022high} compresses images from high-dimensional pixel space into low-dimensional feature space via pretrained autoencoder, and builds diffusion model based on latent feature space such that computational overhead for both training and inference can be dramatically lowered. The training of LDM is similar to Eq. \ref{conditional_DDPM_loss} except that we use notation $z$ to denote latent features:
\begin{equation}
	L=\|\epsilon_{t}-\epsilon_{\theta}(z_{t}, t, c)\|_{2},
	\label{LDM_loss}
\end{equation}
where $\epsilon_{t} \sim \mathcal{N}(0, \mathcal{I})$, $z_{t}=\sqrt{\bar{\alpha}_{t}}z_{0}+\sqrt{1-\bar{\alpha}_{t}}\epsilon_{t}$, $z_{0}=E(x_{0})$, $x_{0}$ is the noise-free clean image, $E$ is the pretrained image encoder. The reverse denoising process from $z_{T} \sim \mathcal{N}(0, \mathcal{I})$ to $z_{0}$ is the same as $x_{T} \sim \mathcal{N}(0, \mathcal{I})$ to $x_{0}$ in DDPM. After reverse denoising process, the denoised clean features $z_{0}$ is decoded by the pretrained decoder $D$ to yield the finally generated image $x_{0}$, \textit{i.e.}, $x_{0}=D(z_{0})$. In LDM framework, the condition signal $c$ could be the extracted image features that are concatenated with $z_{t}$ as the input of $\epsilon_{\theta}$ for image-to-image translation applications, and also could be the encoded textual features that are interacted with $z_{t}$ via cross-attention layers inside $\epsilon_{\theta}$ for text-to-image synthesis.

\subsection{Adversarially Distilled Diffusion Model}
Adversarial diffusion distillation (ADD) \cite{sauer2024adversarial} reduces the inference steps of diffusion model required for high-quality image generation from over 50 steps to just 1–4 steps, enabling near real-time synthesis. The core components of ADD consist of score distillation from a pretrained teacher diffusion model and adversarial training with a discriminator network. The SDXL-Turbo, which we use in our proposed method, is a typical application of ADD that distills knowledge from the pretrained SDXL model. ADD integrates two complementary loss functions:
\begin{equation}
	L=L_{adv}(\hat{x}_{\theta}(x_{t}, t), \phi)+\lambda L_{distill}(\hat{x}_{\theta}(x_{t}, t), \psi),
\end{equation}
where $\theta$ denotes parameters of the ADD student model (initialized from pretrained SDXL), $\psi$ denotes fixed parameters of the pretrained and frozen teacher diffusion model, $\phi$ denotes parameters of the discriminator network, $\lambda$ is the balancing hyperparameter. The $\hat{x}_{\theta}(x_{t}, t)$ is the predicted clean image (or clean latent representation) produced by the ADD student model, parameterized by $\theta$, given the noisy input $x_{t}$ at a specific time step $t$:
\begin{equation}
	\hat{x}_{\theta}(x_{t}, t)=\frac{x_{t}-\sqrt{1-\bar{\alpha}_{t}\epsilon_{\theta}(x_{t}, t)}}{\sqrt{\bar{\alpha}_{t}}},
\end{equation}

where $\epsilon_{\theta}$ is the noise estimation network of the ADD student model. The adversarial loss ensures that the generated images lie directly on the manifold of real images, mitigating the blurriness and artifacts common in distilled models. In adversarial training, the student diffusion model acts as the generator, the discriminator comprises a frozen pretrained feature backbone $F$ (i.e., DINOv2 ViT-L) and a learnable lightweight discriminator heads $D_{\phi,k}$ attached to different layers $k$ of the feature network $F$. Optional conditions such as text embeddings $c_{text}$ and image embeddings $c_{img}$ can be injected into both the
generator and the discriminator. 

Using the Hinge Loss objective, the generator adversarial loss is formulated as:
\begin{equation}
	L_{adv}^{G}(\hat{x}_{\theta}(x_{t}, t), \phi)=-\mathbb{E}_{t,\epsilon,x_{0}}[\sum_{k}D_{\phi,k}(F_{k}(\hat{x}_{\theta}(x_{t}, t)))].
\end{equation}
The corresponding discriminator loss is formulated as:
\begin{align}
	L_{adv}^{D}(\hat{x}_{\theta}(x_{t}, t), \phi) = & \mathbb{E}_{x_{0}} \left[\sum_{k} \max(0, 1-D_{\phi, k}(F_{k}(x_{0}))) + \gamma R1(\phi)\right] \nonumber \\
	 + & \mathbb{E}_{\hat{x}_{\theta}} \left[\sum_{k} \max(0, 1+D_{\phi, k}(F_{k}(\hat{x}_{\theta})))\right],
\end{align}
where the first term encourages positive outputs for real images $x_{0}$, the second term encourages negative outputs for generated images $\hat{x}_{0}$. $R1(\phi)$ denotes the $R1$ gradient penalty, which stabilizes training by penalizing the gradient norm of discriminator outputs with respect to real data:
\begin{equation}
	R1(\phi)=\frac{1}{2}\mathbb{E}_{x_{0}}[||\nabla_{x_{0}}D_{\phi}(x_{0})||^{2}].
\end{equation}
The distillation loss transfers knowledge from the teacher diffusion model to the student model. Given the image $\hat{x}_{\theta}$ generated by the ADD student model, ADD firstly perturbs the generated $\hat{x}_{\theta}$ through teacher's forward diffusion process, yielding a noisy image at a randomly sampled time step $t$:
\begin{equation}
	\hat{x}_{\theta,t}=\sqrt{\bar{\alpha}_{t}}x_{0}+\sqrt{1-\bar{\alpha}_{t}}\epsilon_{t}, \ \ \epsilon_{t} \sim \mathcal{N}(0, \mathcal{I}).
\end{equation}
Then, the teacher model performs denoising prediction on the perturbed sample $\hat{x}_{\theta,t}$, and compute distance between student output and teacher reconstruction, which is formulated as:
\begin{equation}
	L_{distill}(\hat{x}_{\theta}(x_{t}, t), \psi)=\mathbb{E}_{t, \epsilon_{t}}[c(t)\cdot d(\hat{x}_{\theta}, \hat{x}_{\psi}(sg(\hat{x}_{\theta,t});t))],
\end{equation}
where $sg(\cdot)$ denotes the stop-gradient operation (teacher gradients do not backpropagate to student), $d(\cdot, \cdot)$ is a distance metric such as MSE, $c(t)$ is a time-dependent weighting coefficient. The teacher's reconstruction is defined as:
\begin{equation}
	\hat{x}_{\psi}(\hat{x}_{\theta, t}, t)=\frac{\hat{x}_{\theta, t}-\sqrt{1-\bar{\alpha}_{t}\hat{\epsilon}_{\psi}(\hat{x}_{\theta, t}, t)}}{\sqrt{\bar{\alpha}_{t}}},
\end{equation}
where $\hat{\epsilon}_{\psi}(\hat{x}_{\theta, t}, t)$ is the teacher's predicted noise.

\subsection{Classifier-free Guidance}
Classifier-free guidance (CFG) \cite{ho2021classifier} is a technique used in conditional generative models, particularly diffusion probabilistic models, to control the trade-off between sample diversity and fidelity to the conditioning information (e.g., a text prompt or a class label). Unlike classifier guidance, which requires a separately trained classifier to steer the generation process, CFG achieves this control without any auxiliary model, relying solely on the original generative model's own estimates of conditional and unconditional score functions. In CFG, the predicted noise $\epsilon_{t}$ at time step $t$ interpolates between the text-conditioned and unconditional noise estimations: 
\begin{equation}
	\epsilon_{t}=\omega \epsilon_{\theta}(x_{t}, t, y) + (1-\omega)\epsilon_{\theta}(x_{t}, t, \emptyset),
\end{equation}
where $\emptyset$ denotes the null text embedding which is encoded from the empty string ``'', $\omega$ is the guidance strength parameter controlling the intensity of CFG. For $\omega$=0, the generation is purely unconditional, ignoring the condition entirely. For $\omega$=1, the standard conditional model is recovered. For $\omega$>1, the condition is artificially amplified, forcing the sample to adhere more strictly to the provided context.

\subsection{DCT Background}
The Discrete Cosine Transform (DCT) is a fundamental technique in signal processing and data compression. It belongs to a family of unitary transforms that express a finite sequence of data points as a sum of cosine functions oscillating at different frequencies. Unlike the Discrete Fourier Transform (DFT), the DCT operates purely on real numbers and avoids the discontinuities at block boundaries by implicitly assuming symmetric extension of the signal. In this method, the DCT is employed to extract low-frequency components, offering the advantage of quantitatively controlling the extent of low-frequency signal extraction by adjusting the filtering threshold. We use the DCT-II form where the one-dimensional DCT (1D-DCT) is formulated as:

\begin{equation}
	S[k]=\sqrt{\frac{2}{N}}\cdot c[k]\cdot \sum_{n=0}^{N-1}x[n]\cdot \cos(\frac{\pi k(2n+1)}{2N}),
	\label{eq:1D-DCT}
\end{equation}
\begin{equation}
	c[k] = 
	\begin{cases} 
		\frac{1}{\sqrt{2}} & k = 0 \\
		1 & k \neq 0
	\end{cases}
\end{equation}
where $N$ denotes the sequence length, $k = 0, 1, \dots, N-1$. $x$ refers to the input sequence, $S$ is the corresponding DCT spectrum. 

\section{More Implementation Details}
\subsection{DCT Filtering in Null-text Structure Alignment}
For the DCT filtering operations used in the VLC loss in our method, the input sequence $x$ in Eq. \ref{eq:1D-DCT} is either $z_{t\rightarrow 0}^{(1)}$, the predicted clean latent in the source view, or $g^{-1}(z_{t\rightarrow 0}^{(2)})$, the inverse transformation of the predicted clean latent in the transformed view. Let $H$, $W$ respectively denote the height and width of the latent feature, then the sequence length $N$ in Eq. \ref{eq:1D-DCT} is either $H$ or $W$ depending on which spatial dimension to apply 1D-DCT on. In our method, $dct_{th}^{(h)}(\cdot)$ refers to the operation of applying 1D-DCT transformation over the $h$ dimension of the latent feature followed by the low-pass filtering which removes the frequency components higher than the percentile threshold $th$ in the $h$ dimension. Similarly, $dct_{th}^{(w)}(\cdot)$ denotes the same DCT filtering operation with the 1D-DCT transformation applied over the $w$ dimension of the latent feature. 

\begin{figure}[H]
	\centering
	\includegraphics[width=\linewidth]{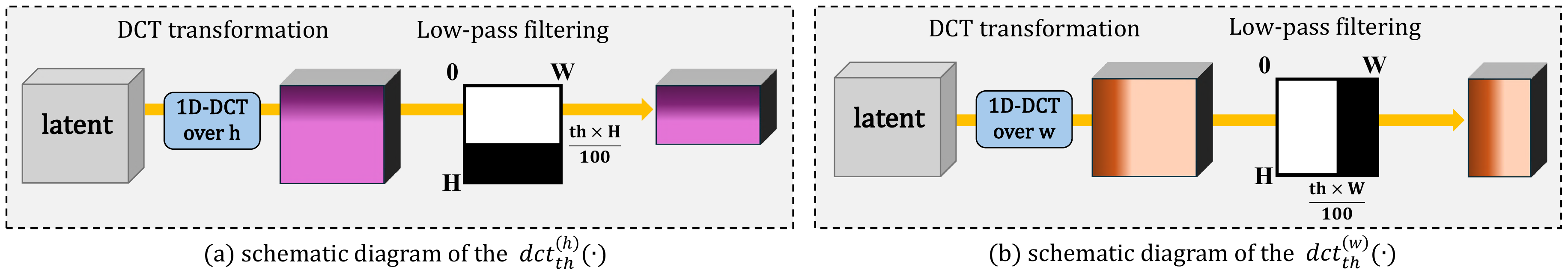}
	\caption{Schematic diagram of $dct_{th}^{(h)}(\cdot)$ and $dct_{th}^{(w)}(\cdot)$ operations applied on the latent feature.}
	\label{fig:DCT_illustration}
\end{figure}

Fig. \ref{fig:DCT_illustration} illustrates the implementation details of the percentile-based low-pass DCT filtering operations, denoted as $dct_{th}^{(h)}(\cdot)$ and $dct_{th}^{(w)}(\cdot)$, which are employed in our proposed VLC loss. For $dct_{th}^{(h)}(\cdot)$, the 1D-DCT is firstly applied along the $h$ dimension of the input latent feature. Subsequently, frequency components exceeding $\frac{th \times H}{100}$ in the $h$ dimension are filtered out, thereby retaining only low-frequency spectral components along the $h$ dimension. Similarly, $dct_{th}^{(w)}(\cdot)$ applies the 1D-DCT along the $w$ dimension of the input latent feature, then removes frequency components higher than $\frac{th \times W}{100}$ in the $w$ dimension, and outputs the resulting low-frequency spectral components.

\subsection{Attention Maps Extraction in Semantic Enhancement}
In cross-modal semantic enhancement optimization, cross-attention maps are extracted and utilized to guide the optimization of input latent feature to enhance semantic expressiveness of the generated results. The computation of the cross-attention maps is reformulated below:
\begin{equation}
	M_{l}^{(1)}=Softmax\left(Q_{l}^{(1)} (K_{l}^{(1)})^{T}/ \sqrt{d}\right), \ \ \
	M_{l}^{(2)}=Softmax\left(Q_{l}^{(2)} (K_{l}^{(2)})^{T} / \sqrt{d}\right),
\end{equation}
where $l$ denotes layer index, $M_{l}^{(1)}$ and $M_{l}^{(2)}$ denote attention maps at the $l$-th cross-attention layer inside the denoising network $\epsilon_{\theta}$, given $z_{t}$ and $g(z_{t})$ as input, respectively, $Q_{l}^{(1)}$ and $Q_{l}^{(2)}$ are queries of the $l$-th cross-attention layer derived from $z_{t}$ and $g(z_{t})$, respectively, $K_{l}^{(1)}$ and $K_{l}^{(2)}$ are keys of the $l$-th cross-attention layer originated from the textual embedding $y_{1}$ and $y_{2}$, respectively, $d$ denotes the feature dimension of the query and key at the current layer. $M_{l}^{(1)}$ comprises multiple attention maps with each attention map associated with a word embedding in $y_{1}$, and similarly for $M_{l}^{(2)}$ with $y_{2}$. We sum over the selected attention maps corresponding to the conceptually most crucial words in $P_{1}$ and $P_{2}$ (marked in red in Fig. \ref{fig:method}), yielding $\tilde{M}_{l}^{(1)}$ and $\tilde{M}_{l}^{(2)}$. Let $L$ denotes the total number of cross-attention layers inside $\epsilon_{\theta}$. By upscaling all the extracted $\{\tilde{M}_{l}^{(1)}\}_{l=1}^{L}$ to the same spatial size of latent features and summing them together, we obtain $\tilde{M}^{(1)}$. Analogously, $\tilde{M}^{(2)}$ is obtained by upscaling all the extracted $\{\tilde{M}_{l}^{(2)}\}_{l=1}^{L}$ to the latent feature resolution and summing them together. In $\tilde{M}^{(1)}$, pixel intensity reflects conceptual correlation strength of each spatial location to the key concept of $P_{1}$, and similarly for $\tilde{M}^{(2)}$ to $P_{2}$. The resulting attention maps $\tilde{M}^{(1)}$ and $\tilde{M}^{(2)}$ are utilized for cross-modal semantic enhancement optimization as formulated by Eq. \ref{eq:se_optimize}.

\section{Evaluation Metrics and Criteria}
\subsection{Evaluation Metrics}
This subsection provides a detailed explanation of the quantitative evaluation metrics used in the main text to assess model performance in generating visual anagram images, along with their corresponding mathematical definitions. Let $P_{1}$ and $P_{2}$ denote the two multi-view text prompts, $x^{(1)}$ and $x^{(2)}$ denote the two views of the generated anagram image that are respectively aligned with $P_{1}$ and $P_{2}$ in semantic concept. The evaluation metrics are defined as follows: \newline
\noindent \textbf{Worst Alignment Score ($\mathcal{A}_{min}$)}: the worst alignment (CLIP image-text cosine similarity) between ($P_{1}$, $x^{(1)}$) and ($P_{2}$, $x^{(2)}$), which is mathematically defined as:
\begin{equation}
	\mathcal{A}_{min}=min[CLIP(P_{1}, x^{(1)}), CLIP(P_{2},x^{(2)})],
\end{equation}
\begin{equation}
	CLIP(P, x)=\frac{E_{text}(P) \cdot E_{img}(x)}{||E_{text}(P)||\ ||E_{img}(x)||},
\end{equation}
where $E_{text}$ and $E_{img}$ are CLIP text encoder and CLIP image encoder, respectively. \newline
\noindent \textbf{Average Alignment Score ($\mathcal{A}_{avg}$)}: the average image-text semantic alignment score across the two views, which is mathematically defined as: 
\begin{equation}
	\mathcal{A}_{avg}=\frac{CLIP(P_{1}, x^{(1)}) + CLIP(P_{2},x^{(2)})}{2}.
\end{equation}
\textbf{Concealment Score ($\mathcal{C}$)}: measuring how well the generated anagram image from one view conceals textual concepts from the other view. This metric evaluates model's performance in suppressing semantic ambiguity, i.e., the ability of the model in presenting semantic alignment of a view to its paired text prompt while mitigating semantic leakage from the other text prompt. The mathematical definition is provided as follows:
\begin{equation}
	u_{11}, u_{12} = Softmax\left[CLIP(P_{1}, x^{(1)}), CLIP(P_{1}, x^{(2)})\right],
\end{equation}
\begin{equation}
	u_{21}, u_{22} = Softmax\left[CLIP(P_{2}, x^{(1)}), CLIP(P_{2}, x^{(2)})\right],
\end{equation}
\begin{equation}
	v_{11}, v_{21} = Softmax\left[CLIP(P_{1}, x^{(1)}), CLIP(P_{2}, x^{(1)})\right],
\end{equation}
\begin{equation}
	v_{12}, v_{22} = Softmax\left[CLIP(P_{1}, x^{(2)}), CLIP(P_{2}, x^{(2)})\right],
\end{equation}
\begin{equation}
	C=(u_{11} + u_{22} + v_{11} + v_{22}) / 4.
\end{equation}

\subsection{Evaluation Criteria in User Study}
The user study shown in Fig. \ref{fig:user_score} is designed to subjectively evaluate methods in generating anagram images. Over 40 participants were recruited to evaluate the output visual anagrams of different methods using a 1–10 rating scale across the following three dimensions:

\begin{enumerate}
	\item \textbf{Visual harmony}: the degree to which the multi-view contents are harmoniously blended together without noticeable structural artifacts and unnatural visual elements;
	\item \textbf{Conceptual fidelity}: the faithfulness of each view to its corresponding text prompt in semantic concept, along with the ability to conceal concept from the other prompt;
	\item \textbf{Semantic saliency}: the degree of clarity and distinctiveness of the semantic content in each view, where semantically blurry or ambiguous results are considered less preferable.
\end{enumerate}

\section{Auxiliary Experiments}
\subsection{Ablation Study on Denoising Steps}
\begin{table}[htbp]
	\centering
	\caption{Ablation study on the denoising steps $T$ on the CIFAR10 2-views dataset.}
	\label{tab:ablation_T}
	\begin{tabular*}{\columnwidth}{>{\centering\arraybackslash}p{0.2\columnwidth}@{\extracolsep{\fill}} |cccc}
		\hline
		denoising steps $T$ & $\mathcal{A}_{min}\uparrow$ & $\mathcal{C}\uparrow$ & $\mathcal{A}_{avg}\uparrow$ & $\mathcal{AS}\uparrow$\\
		\hline
		1   & 0.2486 & 0.6702 & 0.2618 & 3.55\\
		2   & 0.2579 & 0.6887 & 0.2704 & 4.73\\
		3   & 0.2708 & 0.6915 & 0.2818 & 6.05\\
		4   & \textbf{0.2726} & \textbf{0.6918} & \textbf{0.2824} & \textbf{6.14}\\
		5   & 0.2675 & 0.6876 & 0.2800 & 5.86\\
		6   & 0.2603 & 0.6851 & 0.2715 & 5.44\\
		\hline
	\end{tabular*}
\end{table}
Table \ref{tab:ablation_T} presents ablation study on the total number of the denoising steps $T$ in our method. Results show that the textual fidelity and the aesthetic score of the generated anagram images improve with the increase of $T$, but drop rapidly when $T>4$. This is due to that our model is built upon SDXL-Turbo, which achieves optimal generation quality within $1 \leq T \leq 4$. Extending $T$ to $T>4$ degrades visual quality of the generated images and thus leads to declined evaluation metrics. Within $1 \leq T \leq 4$, smaller value of $T$ results in worse evaluation results, suggesting that the multi-view synchronous denoising is not well-suited for very few sampling steps (e.g., $T=1$ or $T=2$). This conclusion aligns with our observation that our method with $T=1$ or $T=2$ produces overly blurred anagram images which struggle to sufficiently reflect the target semantic concepts.

\subsection{Ablation Study on Inner Loop Steps}
\begin{table}[htbp]
	\centering
	\caption{Ablation study on the inner loop steps $K$ on the CIFAR10 2-views dataset.}
	\label{tab:ablation_K}
	\begin{tabular*}{\columnwidth}{>{\centering\arraybackslash}p{0.2\columnwidth}@{\extracolsep{\fill}}|>{\centering\arraybackslash}c >{\centering\arraybackslash}c >{\centering\arraybackslash}c|>{\centering\arraybackslash}c}
		\hline
		inner loop step $K$ & $\mathcal{A}_{min}\uparrow$ & $\mathcal{C}\uparrow$ & $\mathcal{A}_{avg}\uparrow$ & inference time \\
		\hline
		0   & 0.2482 & 0.6690 & 0.2677 & 0.4s \\
		1   & 0.2544 & 0.6736 & 0.2706 & 0.9s \\
		2   & 0.2617 & 0.6821 & 0.2739 & 1.4s \\
		3   & 0.2668 & 0.6874 & 0.2773 & 1.8s \\
		4   & 0.2707 & 0.6905 & 0.2809 & 2.2s \\
		5   & 0.2726 & 0.6918 & 0.2824 & 2.6s \\
		6   & 0.2733 & 0.6923 & 0.2827 & 3.1s \\
		7   & 0.2738 & 0.6925 & 0.2829 & 3.6s \\
		8   & 0.2740 & 0.6926 & 0.2829 & 4.1s \\
		\hline
	\end{tabular*}
\end{table}
We also perform ablation study on the inner loop steps $K$ in our structure-semantic co-optimization (S2CO) framework. Results presented in Tab. \ref{tab:ablation_K} demonstrate that increasing the inner loop steps $K$ (the number of S2CO optimization times at each denoising step) yields monotonic performance gains in multi-view textual fidelity, although the improvements become marginal for $K>5$. Results also show that the overall inference time scales approximately linearly with the increase of $K$. Through a balanced consideration of the quantitative metrics and the time cost, we select $K=5$ as the optimal trade-off between model performance and computational efficiency.

\subsection{Quantitative Verification of Structural Divergence}
We design the following experiment to quantitatively verify the structural divergence issue in the multi-view synchronous denoising process and how the null-text structure alignment (NTSA) mitigates this issue. Based on a standard $T$-step multi-view parallel denoising trajectory composed of $z_{T}\rightarrow \cdots z_{t} \rightarrow \cdots z_{1} \rightarrow z_{0}$ (no null-text structure alignment and semantic enhancement optimizations involved), we apply attention-guided noise fusion at the first $T-1$ steps. For the last denoising step $z_{1} \rightarrow z_{0}$, we do not perform multi-view noise fusion and instead directly predict the two image views $x_{0}^{(1)}$ and $x_{0}^{(2)}$ according to their respective noise estimations:
\begin{equation}
	\epsilon_{1}^{(1)}=\omega\epsilon_{\theta}(z_{1}, t=1, y_{1})+(1-\omega)\epsilon_{\theta}(z_{1}, t=1, y_{\emptyset}), 
\end{equation}
\begin{equation}
	\epsilon_{1}^{(2)}=\omega\epsilon_{\theta}(g(z_{1}), t=1, y_{2})+(1-\omega)\epsilon_{\theta}(g(z_{1}), t=1, y_{\emptyset}),
\end{equation}
\begin{equation}
	z_{0}^{(1)}=(z_{1}-\sqrt{1-\bar{\alpha}_{1}}\epsilon_{1}^{(1)})/\sqrt{\bar{\alpha}_{1}}, \ \  z_{0}^{(2)}=(g(z_{1})-\sqrt{1-\bar{\alpha}_{1}}\epsilon_{1}^{(2)})/\sqrt{\bar{\alpha}_{1}},
\end{equation}
\begin{equation}
	x_{0}^{(1)}=D(z_{0}^{(1)}), \ \ \ x_{0}^{(2)}=D(z_{0}^{(2)}),
\end{equation}
where $D$ is VAE decoder, $y_{1}$ and $y_{2}$ are textual embeddings encoded from the multi-view text prompts $P_{1}$ and $P_{2}$, respectively. Structural divergence occurs between $x_{0}^{(1)}$ and $g^{-1}(x_{0}^{(2)})$ due to different semantic guidance from $y_{1}$ and $y_{2}$. We use the \textbf{DINO-ViT Self-Similarity Distance} \cite{chung2022improving} as the metric to measure the structure distance between $x_{0}^{(1)}$ and $g^{-1}(x_{0}^{(2)})$, which we denote as \textbf{Structure Distance w/o NTSA}. The larger this metric is, the more structural divergence between the generated two images when transformed back to the original view. 

To quantitatively verify the effectiveness of NTSA in narrowing structural divergence, we apply NTSA at the final denoising step $z_{1}\rightarrow z_{0}$, with the generated two images recomputed from the optimized null-text embedding $\tilde{y}_{\emptyset}$, where $\tilde{y}_{\emptyset}$ is optimized with $K$ steps under our proposed VLC loss. The generated two images with the last-step NTSA are formulated as:
\begin{equation}
	\tilde{\epsilon}_{1}^{(1)}=\omega\epsilon_{\theta}(z_{1}, t=1, y_{1})+(1-\omega)\epsilon_{\theta}(z_{1}, t=1, \tilde{y}_{\emptyset}), 
\end{equation}
\begin{equation}
	\tilde{\epsilon}_{1}^{(2)}=\omega\epsilon_{\theta}(g(z_{1}), t=1, y_{2})+(1-\omega)\epsilon_{\theta}(g(z_{1}), t=1, \tilde{y}_{\emptyset}),
\end{equation}
\begin{equation}
	\tilde{z}_{0}^{(1)}=(z_{1}-\sqrt{1-\bar{\alpha}_{1}}\tilde{\epsilon}_{1}^{(1)})/\sqrt{\bar{\alpha}_{1}}, \ \  \tilde{z}_{0}^{(2)}=(g(z_{1})-\sqrt{1-\bar{\alpha}_{1}}\tilde{\epsilon}_{1}^{(2)})/\sqrt{\bar{\alpha}_{1}},
\end{equation}
\begin{equation}
	\tilde{x}_{0}^{(1)}=D(\tilde{z}_{0}^{(1)}), \ \ \ \tilde{x}_{0}^{(2)}=D(\tilde{z}_{0}^{(2)}),
\end{equation}
Similarly, we use DINO-ViT Self-Similarity Distance to measure the structure distance between $\tilde{x}_{0}^{(1)}$ and $g^{-1}(\tilde{x}_{0}^{(2)})$, which we denote as \textbf{Structure Distance w/ NTSA}. 

We randomly sample 10 pairs of ($P_{1}$,$P_{2}$) in CIFAR10 2-views dataset, and evaluate the average Structure Distance w/o and w/ NTSA for different denoising steps $T$ based on the above-mentioned experiment setting. Results evaluated with the vertical flipping transformation and the 90\degree \ rotation transformation are respectively reported in Tab. \ref{tab:structure_distance_flipping} and Tab. \ref{tab:structure_distance_rotation}. The non-zero values of the Structure Distance metric indicate that structural divergence indeed exists along the denoising process, and that fewer denoising steps $T$ is more prone to structural divergence issue. By comparing the Structure Distance w/o and w/ NTSA under different denoising steps, we see that NTSA consistently narrows the Structure Distance between the generated two images in the original view, which quantitatively verify the effectiveness of NTSA in mitigating structural divergence.

\begin{table}[htbp]
	\centering
	\caption{Comparison of Structure Distance w/o and w/ NTSA evaluated at different denoising steps in the transformation of vertical flipping. Results are averaged over 10 samples in CIFAR10 2-views.}
	\label{tab:structure_distance_flipping}
	\begin{tabular}{lcccc}
		\toprule
		Metric evaluated in \textbf{vertical flipping transformation} & \(T = 1\) & \(T = 2\) & \(T = 3\) & \(T = 4\) \\
		\midrule
		Structure Distance w/o NTSA & 0.074 & 0.066 & 0.056 & 0.049 \\
		Structure Distance w/ NTSA & 0.046 & 0.040 & 0.035 & 0.032 \\
		\bottomrule
	\end{tabular}
\end{table}

\begin{table}[htbp]
	\centering
	\caption{Comparison of Structure Distance w/o and w/ NTSA evaluated at different denoising steps in the transformation of 90\degree \ rotation. Results are averaged over 10 samples in CIFAR10 2-views.}
	\label{tab:structure_distance_rotation}
	\begin{tabular}{lcccc}
		\toprule
		Metric evaluated in \textbf{90\degree \ rotation transformation} & \(T = 1\) & \(T = 2\) & \(T = 3\) & \(T = 4\) \\
		\midrule
		Structure Distance w/o NTSA & 0.077 & 0.068 & 0.057 & 0.050 \\
		Structure Distance w/ NTSA & 0.049 & 0.042 & 0.036 & 0.034 \\
		\bottomrule
	\end{tabular}
\end{table}

\subsection{Quantitative Investigation of the Percentile Parameter in Semantic Enhancement}
In semantic enhancement optimization, the extracted attention maps $\tilde{M}^{(1)}$ and $\tilde{M}^{(2)}$ are leveraged to optimize the input latent feature at each denoising step, which we reformulate below:
\begin{equation}
	z_{t} := z_{t} + \lambda_{2} \nabla_{z_{t}} \left( \sum_{(i,j) \in \Omega^{(1)}} \tilde{M}^{(1)}_{i,j} + \sum_{(i,j) \in \Omega^{(2)}} \tilde{M}^{(2)}_{i,j} \right), \ \ 
	\Omega^{(k)}=\left\{(i,j)|\tilde{M}^{(k)}_{i,j}\geq \tau_{p}^{(k)}\right\},
\end{equation}
where $i,j$ denote spatial indices, $\tau_{p}^{(k)}$ refers to the $p$-th percentile of the values in $\tilde{M}^{(k)}$. We only back-propagate through locations above the $p$-th percentile of each attention map to reinforce semantic expressiveness only in the most concept-relevant regions. We conduct ablation study over the percentile parameter $p$ on a randomly selected subset of CIFAR10 2-views dataset which contains 20 ($P_{1}$, $P_{2}$) samples. Aesthetic Score ($\mathcal{AS}$) is evaluated over different values of $p$, with the average results reported in Tab. \ref{tab:ablation_p}. Setting $p=100$ is equivalent to the absence of semantic enhancement, which produces results with the worst aesthetic quality. Setting $p=0$ corresponds to the maximum semantic enhancement, which over-amplifies attention activations and causes semantic information to diffuse into conceptually irrelevant image regions, thereby degrading the aesthetic score. Through hyper-parameter tuning, we quantitatively find that $p=60$ yields the best aesthetic quality of the generated anagram images.

\begin{table}[htbp]
	\centering
	\caption{Ablation study over the percentile parameter $p$ in semantic enhancement optimization.}
	\label{tab:ablation_p}
	\begin{tabular*}{0.95\textwidth}{l@{\extracolsep{\fill}}cccccc}
		\toprule
		Metric & $p=0$ & $p=20$ & $p=40$ & $p=60$ & $p=80$ & $p=100$ \\
		\midrule
		Aesthetic Score ($\mathcal{AS}$) & 5.96 & 6.03 & 6.09 & 6.12 & 5.98 & 5.87 \\
		\bottomrule
	\end{tabular*}
\end{table}

\subsection{Evaluation on Other Form of Structure Alignment}
Our focus is not latent-space DCT filtering itself, but rather aligning cross-view structure by optimizing null-text embedding to disentangle structure from semantics——a principle that generalizes beyond DCT; other frequency-domain filtering, e.g., wavelet, could also be applicable (Tab. \ref{tab:different_transformations}):
\begin{table}[htbp]
	\centering
	\caption{Comparison between different frequency-domain transformations for structure alignment.}
	\label{tab:different_transformations}
	\begin{tabular}{lccccc}
		\toprule
		\textbf{Frequency-domain filter for NTSA} & $\mathcal{A}_{min}\uparrow$ & $\mathcal{C}\uparrow$ & $\mathcal{A}_{avg}\uparrow$ & $\mathcal{AS}\uparrow$ & Inference Time $\downarrow$ \\
		\midrule
		DCT filter & 0.2719 & 0.6913 & 0.2821 & 6.16 & 2.2s \\
		Wavelet filter & 0.2726 & 0.6918 & 0.2824 & 6.14 & 2.6s \\
		\bottomrule
	\end{tabular}
\end{table}

\subsection{Evaluation on an LLM-Augmented Prompt Dataset}
To evaluate methods under more challenging semantic conditions, we constructed an additional multi-view text prompt dataset using large language model. We defined four types of paired text prompts templates: (``painting of A'', ``painting of B''), (``oil painting of A'', ``oil painting of B''), (``watercolor painting of A'', ``watercolor painting of B''), and (``artistic drawing of A'', ``artistic drawing of B''). ChatGPT was then employed to randomly complete the entity nouns A and B in the templates, resulting in a text prompt dataset consisting of 120 samples of ($P_{1}$, $P_{2}$) pairs, which we term the LLM-Augmented dataset.

Quantitative evaluations based on the curated LLM-Augmented dataset are displayed in Tab. \ref{tab:evaluation_on_augmented_dataset}. Results show that our S2CO-Anagram still achieves top performance in both text fidelity (quantified by $\mathcal{A}_{min}$, $\mathcal{C}$, and $\mathcal{A}_{avg}$) and visual quality (quantified by $\mathcal{AS}$), which is basically consistent with the conclusion obtained on the CIFAR10 2-views dataset as shown in Tab. \ref{tab:quantitative_evaluations}.

\begin{table}[t]
	\centering
	\setlength{\extrarowheight}{0pt}
	\renewcommand{\arraystretch}{1}
	\setlength{\abovetopsep}{0pt}
	\setlength{\belowbottomsep}{0pt}
	\caption{Quantitative evaluations across different methods based on the LLM-Augmented dataset.}
	\label{tab:evaluation_on_augmented_dataset}
	\begin{tabular}{lcccccc}
		\toprule
		\textbf{Method} & $\mathcal{A}_{min}\uparrow$ & $\mathcal{C}\uparrow$ & $\mathcal{A}_{avg}\uparrow$ & $\mathcal{AS}\uparrow$ & Resolution $\uparrow$ & Inference Time $\downarrow$ \\
		\midrule
		Diffusion Illusions & 0.2515 & 0.6845 & 0.2633 & 4.15 & 256$\times$256 & 1027.3s \\
		SyncTweedies       & 0.2538 & 0.6860 & 0.2664 & 4.09 & 256$\times$256 & 5.9s   \\
		Visual Anagrams    & 0.2573 & 0.6778 & 0.2702 & 5.14 & 256$\times$256 & 4.0s   \\
		Anagram-MTL        & 0.2661 & 0.6882 & 0.2775 & 5.36 & 256$\times$256 & 16.2s   \\
		LookingGlass       & 0.2538 & 0.6833 & 0.2684 & 4.83 & 512$\times$512 & 74.8s   \\
		\textbf{S2CO-Anagram} & \textbf{0.2707} & \textbf{0.6894} & \textbf{0.2796} & \textbf{5.92} & \textbf{512$\times$512} & \textbf{2.6s} \\
		\bottomrule
	\end{tabular}
\end{table}

\section{More Visual Results}
\begin{figure*}[h]
	\centering
	\includegraphics[width=\linewidth]{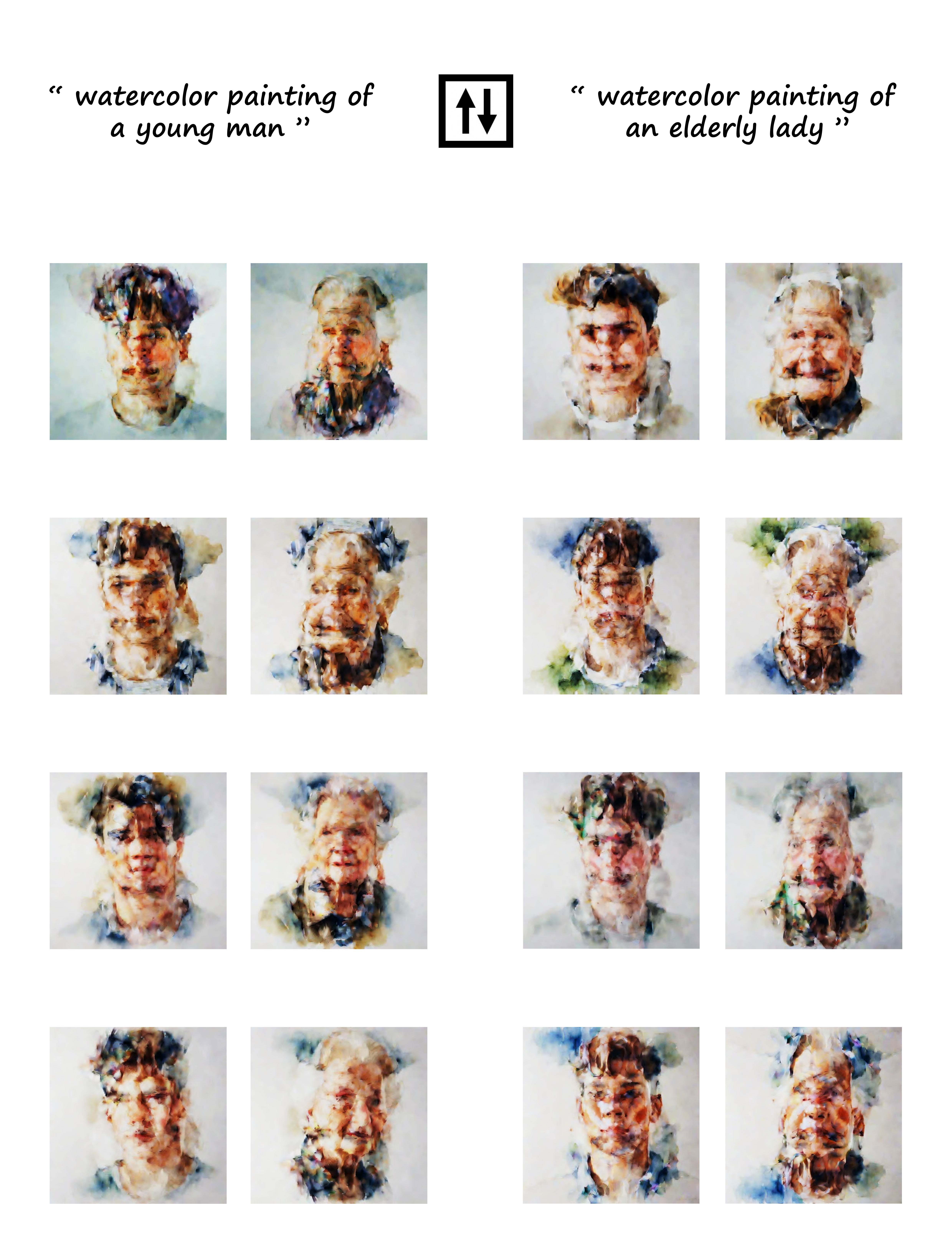}
	\caption{More diverse results-case 1.}
	\label{fig:case1}
\end{figure*}

\begin{figure*}[h]
	\centering
	\includegraphics[width=\linewidth]{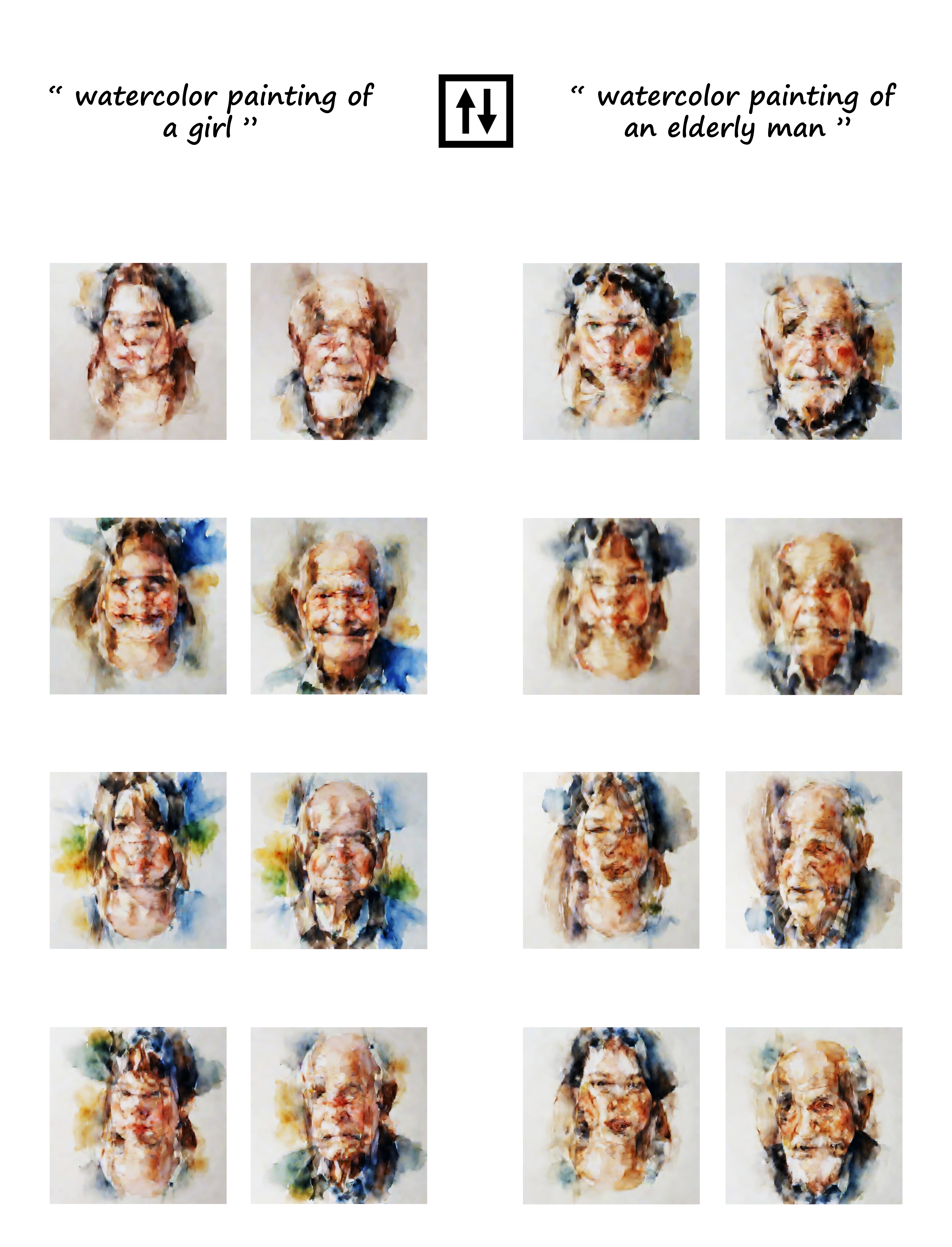}
	\caption{More diverse results-case 2.}
	\label{fig:case2}
\end{figure*}

\begin{figure*}[h]
	\centering
	\includegraphics[width=\linewidth]{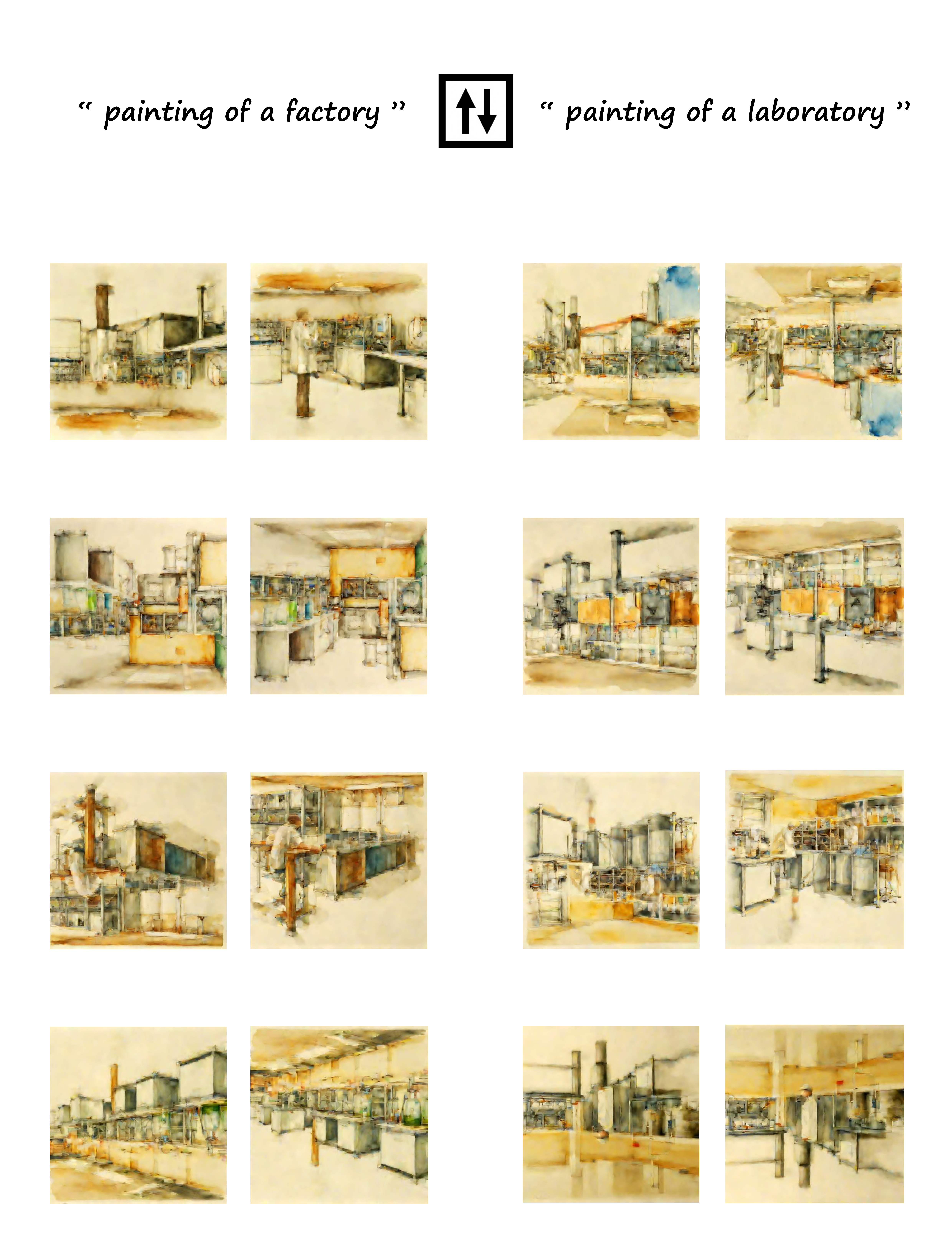}
	\caption{More diversified results-case 3.}
	\label{fig:case3}
\end{figure*}

\begin{figure*}[h]
	\centering
	\includegraphics[width=\linewidth]{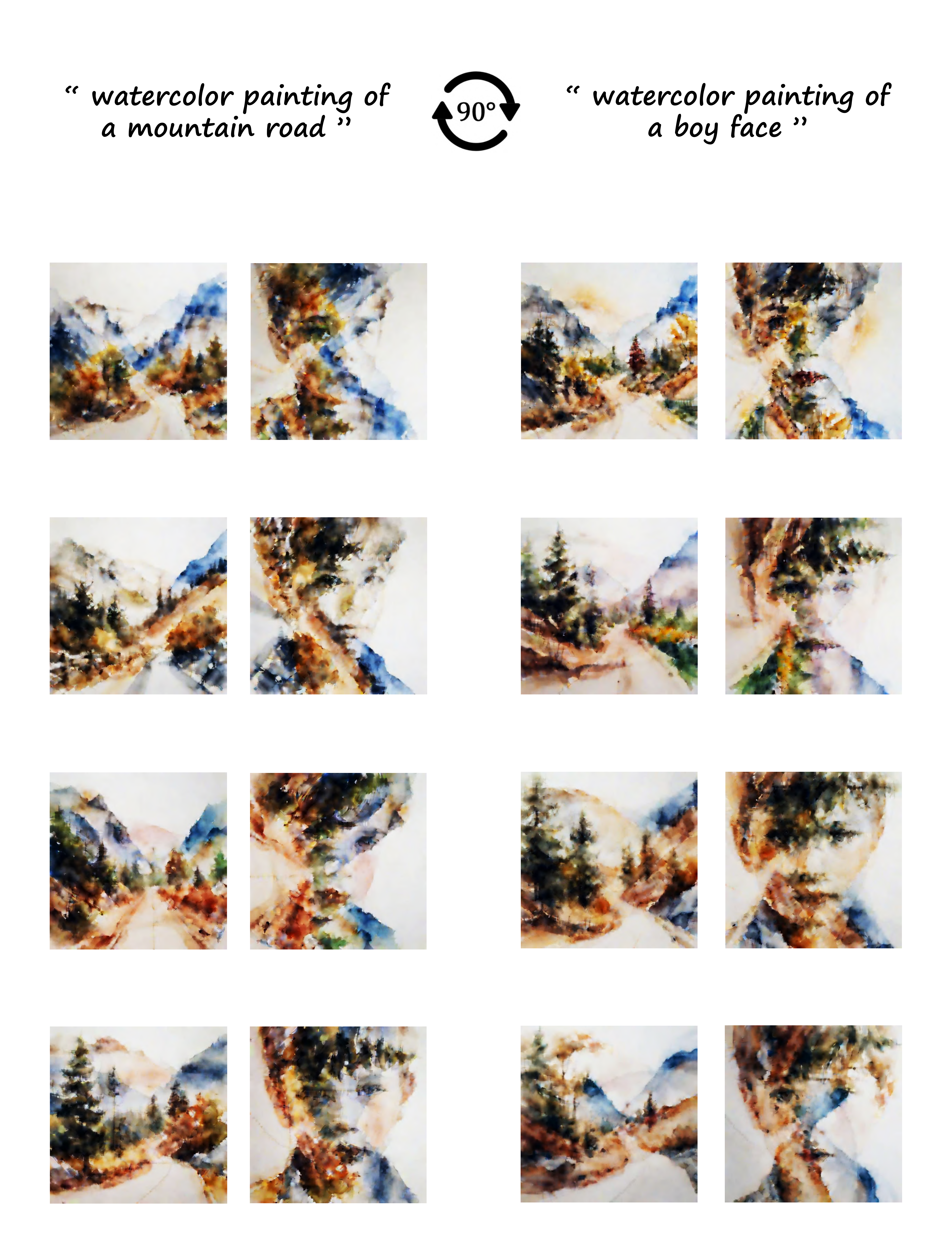}
	\caption{More diverse results-case 4.}
	\label{fig:case4}
\end{figure*}

\begin{figure*}[h]
	\centering
	\includegraphics[width=\linewidth]{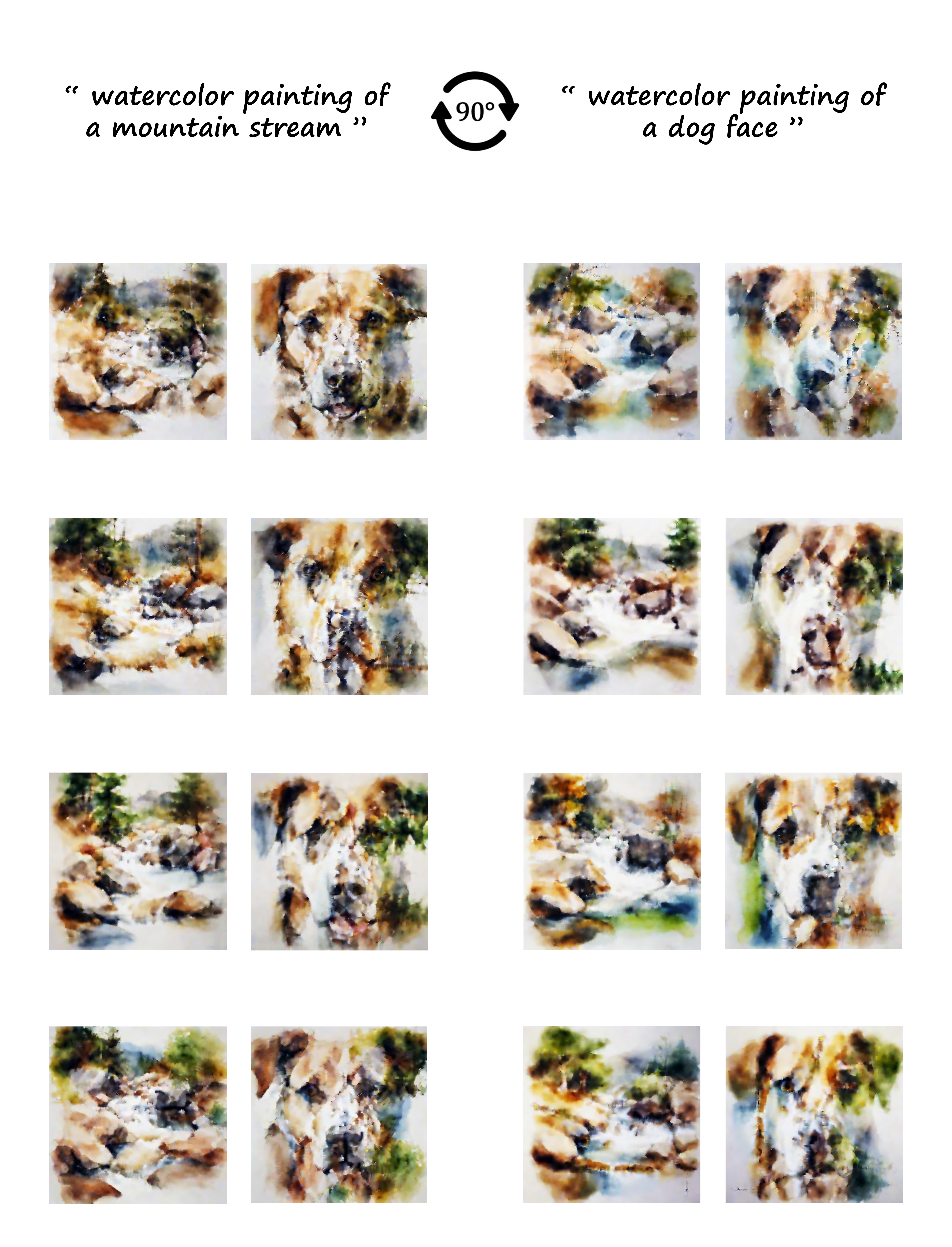}
	\caption{More diversified results-case 5.}
	\label{fig:case5}
\end{figure*}

\begin{figure*}[h]
	\centering
	\includegraphics[width=\linewidth]{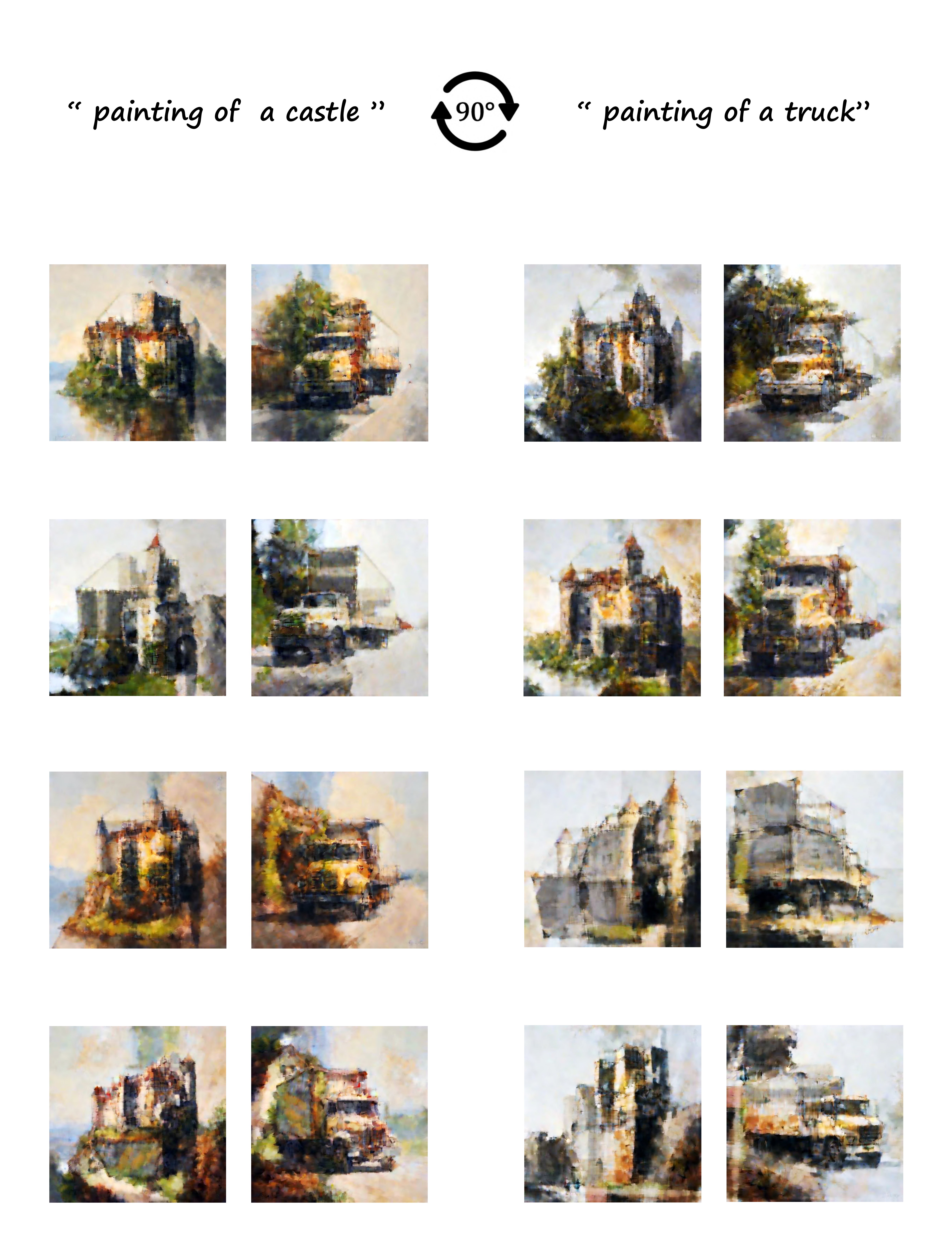}
	\caption{More diversified results-case 6.}
	\label{fig:case6}
\end{figure*}

\begin{figure*}[h]
	\centering
	\includegraphics[width=\linewidth]{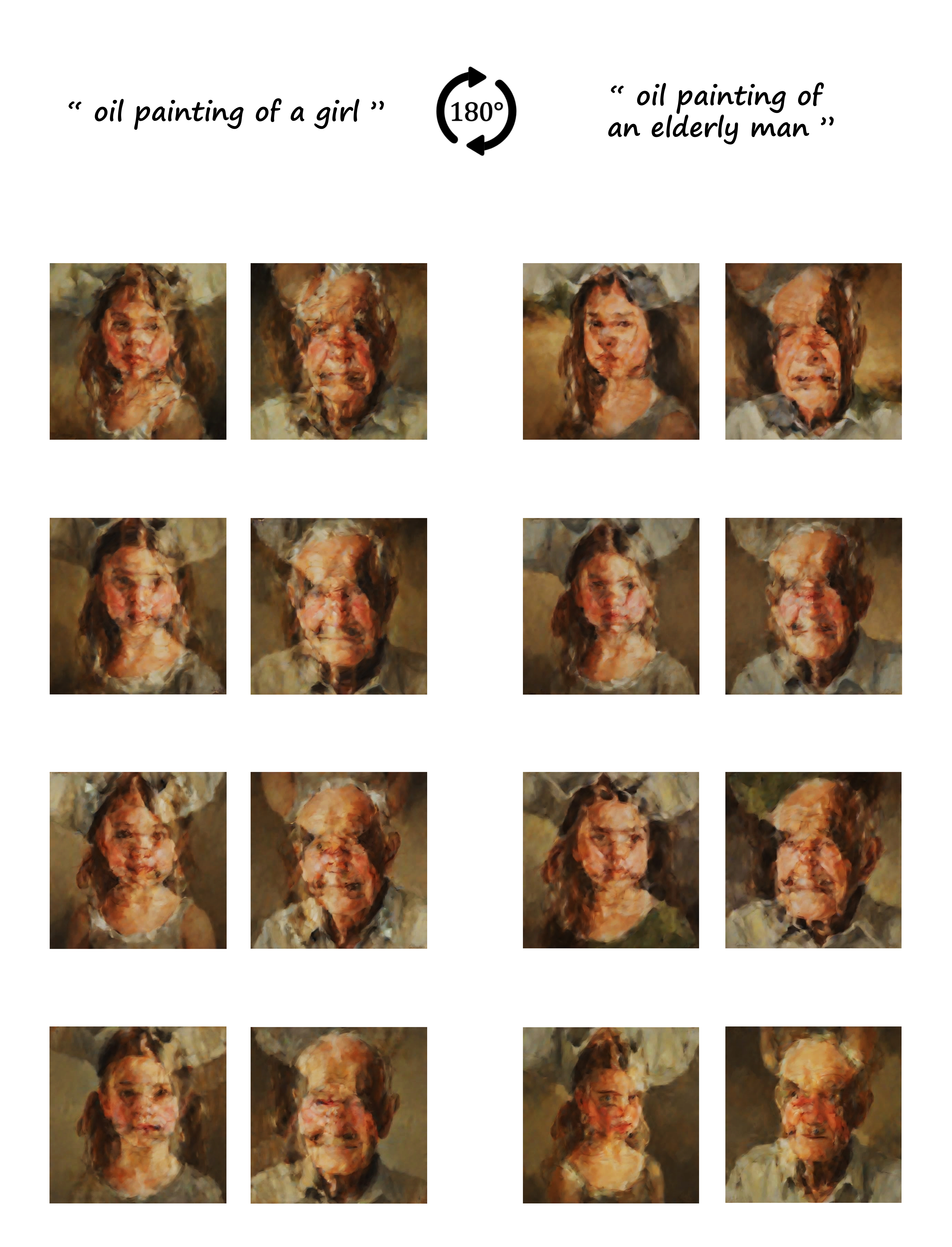}
	\caption{More diversified results-case 7.}
	\label{fig:case7}
\end{figure*}

\begin{figure*}[h]
	\centering
	\includegraphics[width=\linewidth]{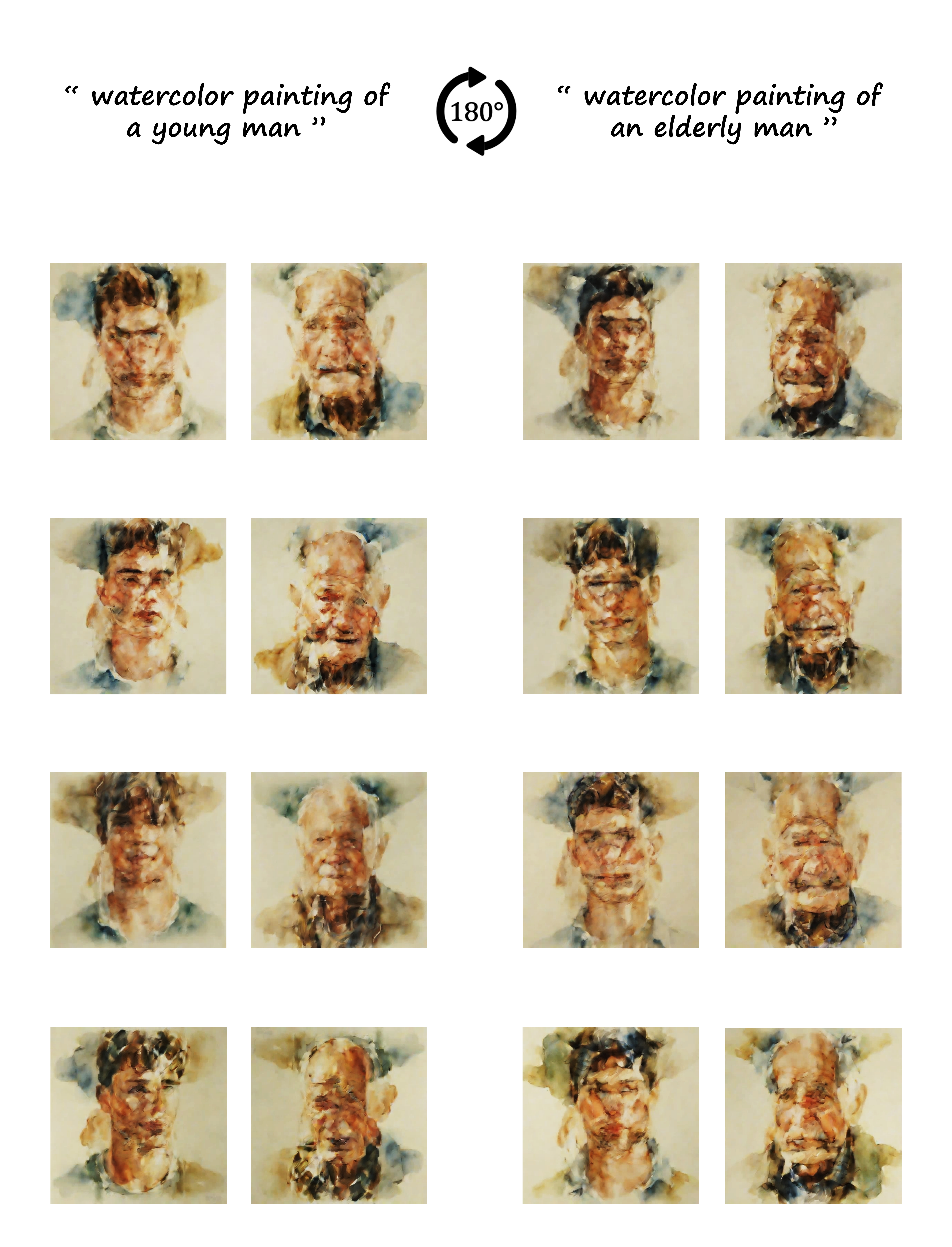}
	\caption{More diversified results-case 8.}
	\label{fig:case8}
\end{figure*}

\begin{figure*}[h]
	\centering
	\includegraphics[width=\linewidth]{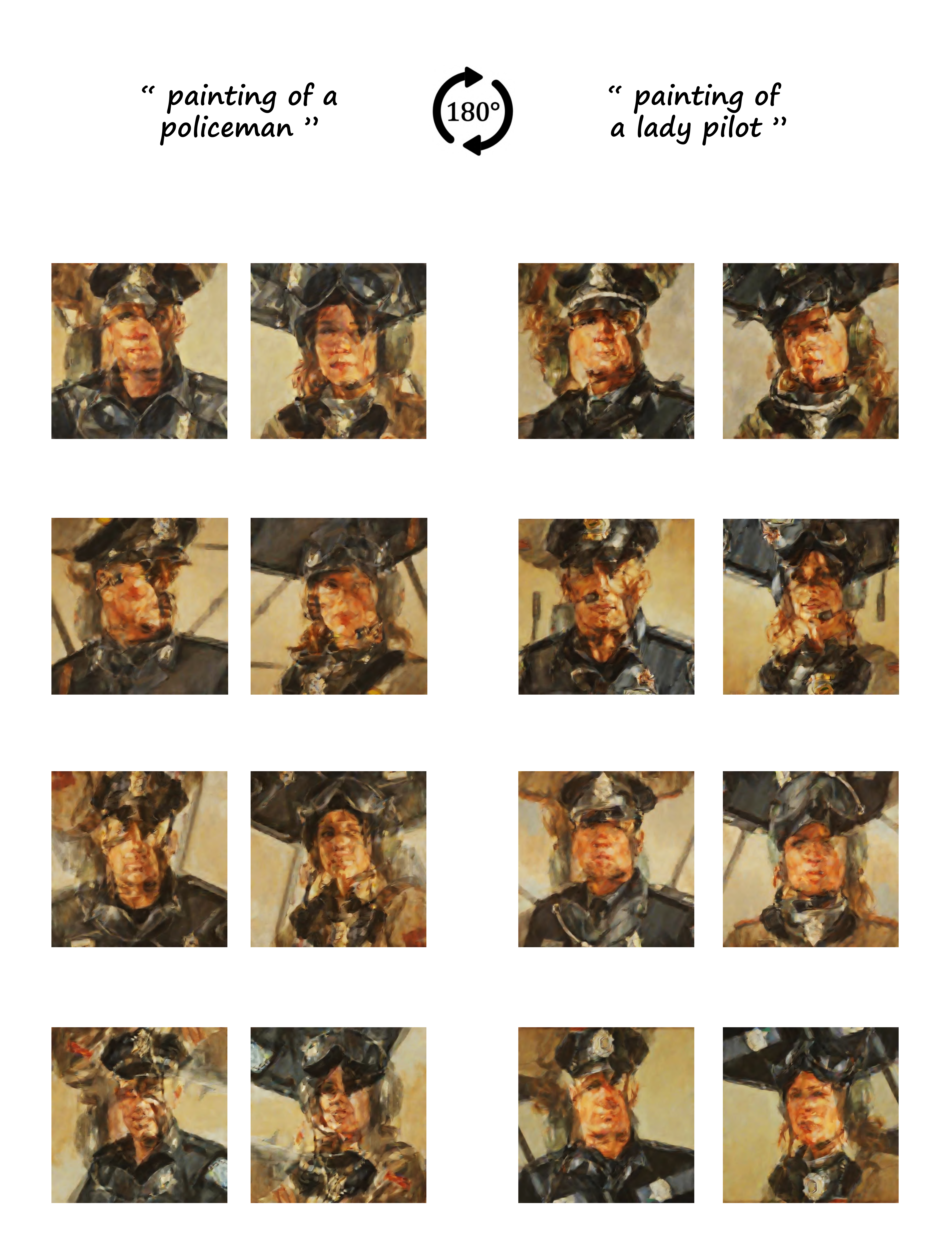}
	\caption{More diversified results-case 9.}
	\label{fig:case9}
\end{figure*}

\begin{figure*}[h]
	\centering
	\includegraphics[width=\linewidth]{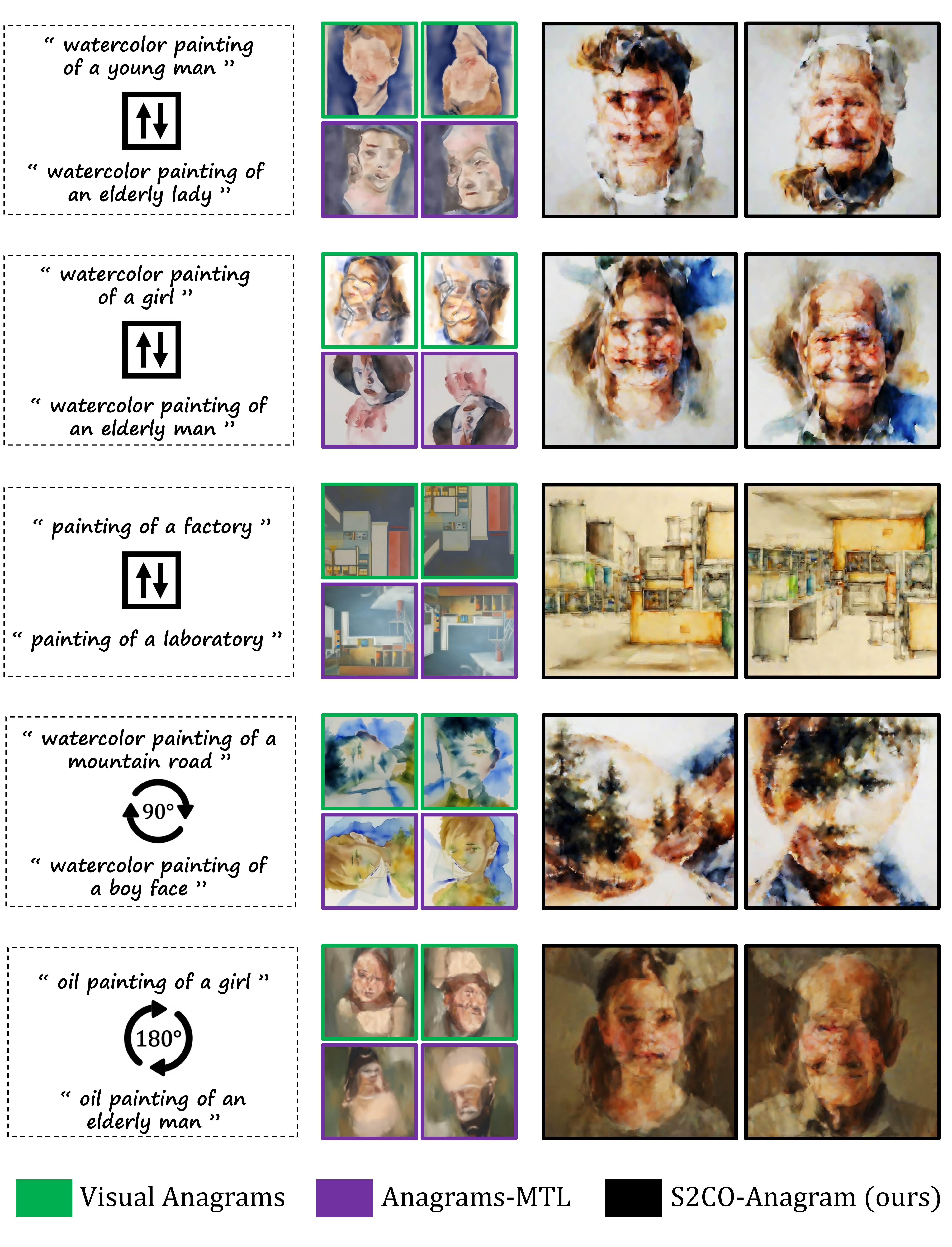}
	\caption{More qualitative method comparison-case 1. Small images are of 256$\times$256 resolution, large images are of 512$\times$512 resolution.}
	\label{fig:case10}
\end{figure*}

\begin{figure*}[h]
	\centering
	\includegraphics[width=\linewidth]{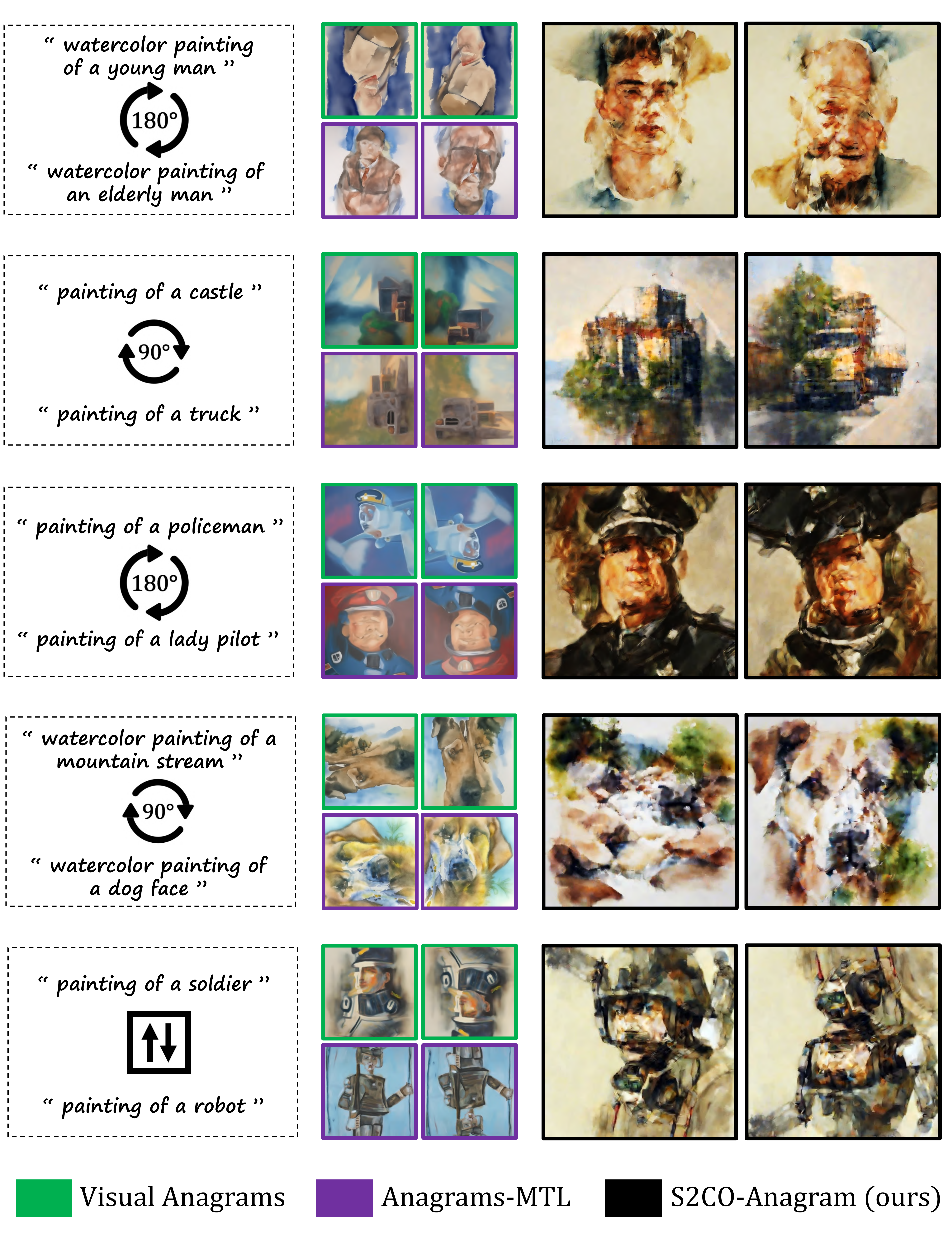}
	\caption{More qualitative method comparison-case 2. Small images are of 256$\times$256 resolution, large images are of 512$\times$512 resolution.}
	\label{fig:case11}
\end{figure*}

To complement visual results of our method, we present additional anagram images generated by our S2CO-Anagram using various transformation functions, as shown in Fig. \ref{fig:case1} through Fig. \ref{fig:case9}. Results fully demonstrate the generative diversity of our method, which produces diverse high-quality samples by randomly initializing the noise latent $z_{T}$ at each runtime. In Fig. \ref{fig:case10} and Fig. \ref{fig:case11}, we display more visual comparison of our method with Visual Anagrams \cite{geng2024visual} and Anagram-MTL \cite{xu2025diffusion}, our results present better image resolution, aesthetic quality, visual harmony, conceptual fidelity, and semantic distinctiveness than the compared approaches.

\section{Limitations and Discussions}
\subsection{Limitations and Future Work}
This paper proposes S2CO-Anagram for high-quality and fast visual anagram synthesis. Our method achieves superior aesthetic quality, text fidelity, semantic clarity, and image resolution than related SOTA approaches with substantially faster inference speed. However, our method still suffer from the following two limitations: (1) our method is capable of generating visually appealing anagram images with an artistic style, however, its performance in generating realistic-style anagram images is relatively poor; (2) we empirically observe that our method fails to generate satisfying results for more complex transformation functions such as jigsaw rearrangement. In future work, we will explore strategies to resolve these issues.
\subsection{Social Impact Discussions}
The proposed S2CO-Anagram framework for fast visual anagram synthesis offers several beneficial societal implications. First, it democratizes illusionary digital art creation by providing artists, designers, and content creators with an efficient tool that generates high-quality visual anagrams in 2 seconds, significantly lowering the barrier to creative expression. Second, educational applications may arise, enabling interactive learning tools that illustrate principles of visual perception, Gestalt psychology, and human creativity. Third, the framework could support therapeutic or cognitive training applications, such as visual rehabilitation exercises or interventions that engage pattern recognition and mental rotation skills. Fourth, this technology could inspire new forms of advertising, branding, and visual communication, in which dual-interpretation imagery captures audience attention and conveys messages in an engaging manner. 

\newpage
\newpage

\newpage

\newpage

\newpage

\newpage

\newpage

\newpage

\newpage

\newpage

\newpage

\newpage

\bibliographystyle{plainnat}   % 或 unsrtnat, abbrvnat 等
\bibliography{references.bib}      % 文件名不加 .bib 后缀
%%%%%%%%%%%%%%%%%%%%%%%%%%%%%%%%%%%%%%%%%%%%%%%%%%%%%%%%%%%%

% \newpage
% \input{checklist.tex}

\end{document}